\documentclass[final]{cvpr}

\usepackage{times}
\usepackage{epsfig}
\usepackage{graphicx}
\usepackage{amsmath}
\usepackage{amssymb}
\usepackage{url}
\usepackage[utf8]{inputenc} 
\usepackage[T1]{fontenc}    
\usepackage{booktabs}       
\usepackage{amsfonts}       
\usepackage{nicefrac}       
\usepackage{microtype}      
\usepackage[title]{appendix}
\usepackage{mathrsfs}
\usepackage[ruled]{algorithm2e}
\usepackage[noend]{algpseudocode}
\usepackage{booktabs}
\usepackage{multirow}
\usepackage{subfigure}
\usepackage{rotating}
\usepackage{xcolor}
\usepackage{caption}
\usepackage{wrapfig}
\usepackage{lipsum}

\usepackage[pagebackref=true,breaklinks=true,colorlinks,bookmarks=false]{hyperref}

\def\cvprPaperID{4340} 
\def\confYear{CVPR 2021}

\begin{document}

\title{Uncertainty-guided Model Generalization to Unseen Domains}

\author{Fengchun Qiao\\
University of Delaware\\
{\tt\small fengchun@udel.edu}
\and
Xi Peng\\
University of Delaware\\
{\tt\small xipeng@udel.edu}
}

\maketitle

\begin{abstract}
We study a worst-case scenario in generalization: Out-of-domain generalization from a single source. The goal is to learn a robust model from a single source and expect it to generalize over many unknown distributions. This challenging problem has been seldom investigated while existing solutions suffer from various limitations. In this paper, we propose a new solution. The key idea is to augment the source capacity in both input and label spaces, while the augmentation is guided by uncertainty assessment. To the best of our knowledge, this is the first work to (1) access the generalization uncertainty from a single source and (2) leverage it to guide both input and label augmentation for robust generalization.
The model training and deployment are effectively organized in a Bayesian meta-learning framework. 
We conduct extensive comparisons and ablation study to validate our approach. The results prove our superior performance in a wide scope of tasks including image classification, semantic segmentation, text classification, and speech recognition.
\end{abstract}

\section{Introduction}

Existing machine learning algorithms have achieved remarkable success under the assumption that training and test data are sampled from similar distributions. When this assumption no longer holds, even strong models ({\it e.g.,} deep neural networks) may fail to produce reliable predictions. 
In this paper, we study a worst-case scenario in generalization: {\it Out-of-domain generalization from a single source.} A model learned from a single source is expected to generalize over a series of unknown distributions. 
This problem is more challenging than {\it domain adaptation}~\cite{motiian2017few,murez2018image,xu2019dsne,liu2019transferable} which usually requires the assessment of target distributions during training, and {\it domain generalization}~\cite{muandet2013domain,ghifary2015domain,li2018learning,carlucci2019jigasaw,dou2019domain} which often assumes the availability of multiple sources.
For example, there exists significant distribution difference in 
medical images collected across different hospitals. The intelligent diagnosis system is required to process images unexplored during training where model update is infeasible due to time or resource limitations.

Recently, \cite{volpi2018generalizing} casts this problem in an ensemble framework. It learns a group of models each of which tackles an unseen test domain.
This is achieved by performing {\it adversarial training}~\cite{goodfellow2014adv} on the source to mimic the unseen test distributions. 
Yet, its generalization capability is limited due to the proposed semantic constraint, which allows only a small amount of data augmentation to avoid semantic changes in the label space.
To address this limitation, \cite{qiao2020learning} proposes {\it adversarial domain augmentation} to relax the constraint. By maximizing the Wasserstein distance between the source and augmentation, the domain transportation is significantly enlarged in the input space.

However, existing data (domain) augmentation based methods~\cite{volpi2018generalizing,peng2018jointly,devries2017cutout,cubuk2019autoaugment,hendrycks2019augmix} merely consider to increase the source capacity by perturbing the input space. Few of them investigate the possibility of label augmentation. An exception is Mixup~\cite{zhang2017mixup} which pioneers label augmentation by randomly interpolating two data examples in both input and label spaces. However, Mixup can hardly address the out-of-domain generalization problem since it is restricted in creating in-domain generations due to the linear interpolation assumption.
Besides, the interpolations are randomly sampled from a fixed distribution, which also largely restricts the flexibility of domain mixtures, yielding sub-optimal performance for unseen domain generalization.  

Another limitation of existing work~\cite{muandet2013domain,ghifary2015domain,li2018learning,carlucci2019jigasaw,dou2019domain} is they usually overlook the potential risk of leveraging augmented data in tackling out-of-domain generalization. This raises serious safety and security concerns in mission-critical applications~\cite{finn2018probabilistic}. For instance, when deploying self-driving cars in unknown environments, it is crucial to be aware of the predictive uncertainty in risk assessment.



To tackle the aforementioned limitations, we propose uncertain out-of-domain generalization.
The key idea is to increase the source capacity guided by uncertainty estimation in both input and label spaces. More specifically, in the input space, instead of directly augmenting raw data~\cite{volpi2018generalizing,qiao2020learning}, we apply uncertainty-guided perturbations to latent features, yielding a domain-knowledge-free solution for various modalities such as image, text, and audio. In the label space, we leverage the uncertainty associated with feature perturbations to augment labels via interpolation, improving generalization over unseen domains. Moreover, we explicitly model the domain uncertainty as a byproduct of feature perturbation and label mixup, guaranteeing fast risk assessment without repeated sampling. Finally, we organize the training and deployment in a Bayesian meta-learning framework that is specially tailored for single source generalization. To summarize, our contribution is multi-fold:
\begin{itemize}
    \item To the best of our knowledge, we are the first to access the uncertainty from a single source. We leverage the uncertainty assessment to gradually improve the domain generalization in a curriculum learning scheme.
    \item For the first time, we propose learnable label mixup in addition to widely used input augmentation, further increasing the domain capacity and reinforcing generalization over unseen domains.
    \item We propose a Bayesian meta-learning method to effectively organize domain augmentation and model training. Bayesian inference is crucial in maximizing the posterior of domain augmentations, such that they can approximate the distribution of unseen domains.
    \item Extensive comparisons and ablation study prove our superior performance in a wide scope of tasks including image classification, semantic segmentation, text classification, and speech recognition.
\end{itemize}

\section{Related Work}

{\bf Out-of-Domain Generalization.}
Domain generalization~\cite{ghifary2015domain,li2017deeper,grubinger2017multi,shankar2018generalizing,carlucci2019jigasaw,dou2019domain} has been intensively studied in recent years. JiGen~\cite{carlucci2019jigasaw} proposed to generate jigsaw puzzles from source domains and leverage them as self-supervised signals.
Wang \textit{et al.}~\cite{wang2020learning} leveraged both extrinsic relationship supervision and intrinsic self-supervision for domain generalization.
Specially, GUD~\cite{volpi2018generalizing} proposed adversarial data augmentation to solve single domain generalization, and learned an ensemble model for stable training. 
M-ADA~\cite{qiao2020learning} extended it to create augmentations with large domain transportation, and designed an efficient meta-learning scheme within a single unified model. Both  GUD~\cite{volpi2018generalizing} and M-ADA~\cite{qiao2020learning} fail to assess the uncertainty of augmentations and only augment the input, while our method explicitly model the uncertainty and leverage it to increase the augmentation capacity in both input and label spaces.
Several methods~\cite{madry2017pgd,wang2019learning,hendrycks2019using} proposed to leverage adversarial training~\cite{goodfellow2014adv} to learn robust models, which can also be applied in single source generalization. PAR~\cite{wang2019learning} proposed to learn robust global representations by penalizing the predictive power of local representations.  
\cite{hendrycks2019using} applied self-supervised learning to improve the model robustness.

{\bf Adversarial training.} 
Szegedy \textit{et al.}~\cite{szegedy2014intriguing} discovered the intriguing weakness of deep neural networks to minor adversarial perturbations.
Goodfellow \textit{et al.}~\cite{goodfellow2014adv} proposed adversarial training to improve model robustness against adversarial samples.
Madry \textit{et al.}~\cite{madry2017pgd} illustrated that adversarial samples generated through 
projected gradient descent can provide robustness guarantees. 
Sinha \textit{et al.}~\cite{sinha2017certifying} proposed principled adversarial training with robustness guarantees through distributionally robust optimization. 
More recently, Stutz \textit{et al.}~\cite{stutz2019disentangling} illustrated that on-manifold adversarial samples can improve generalization. Therefore, models with both robustness and generalization can be achieved at the same time. 
In our work, we leverage adversarial training to create feature perturbations for domain augmentation instead of directly perturbing raw data.


{\bf Meta-learning.}
Meta-learning~\cite{schmidhuber1987evolutionary,thrun2012learning} is a long standing topic on learning models to generalize over a distribution of tasks. 
Model-Agnostic Meta-Learning (MAML)~\cite{finn2017model} is a recent gradient-based method for fast adaptation to new tasks.
In this paper, we propose a modified MAML to make the model generalize over the distribution of domain augmentation.
Several approaches~\cite{li2018learning,balaji2018metareg,dou2019domain} have been proposed to learn domain generalization in a meta-learning framework. 
Li \textit{et al.}~\cite{li2018learning} firstly applied MAML in domain generalization by adopting an episodic training paradigm. 
Balaji \textit{et al.}~\cite{balaji2018metareg} proposed to meta-learn a regularization function to train networks which can be easily generalized to different domains. Dou \textit{et al.}~\cite{dou2019domain} incorporated global and local constraints for learning semantic feature spaces in a meta-learning framework.
However, these methods cannot be directly applied for single source generalization since there is only one distribution available during training.

{\bf Uncertainty Assessment.}
Bayesian neural networks~\cite{hinton1993keeping,graves2011practical,blundell2015weight} have been intensively studied to integrate uncertainty into weights of deep networks. Instead, we apply Bayesian inference to assess the uncertainty of domain augmentations.
Several Bayesian meta-learning frameworks~\cite{grant2018recasting,finn2018probabilistic,yoon2018bayesian,lee2019learning} have been proposed to model the uncertainty of few-shot tasks. 
Grant \textit{et al.}~\cite{grant2018recasting} proposed the first Bayesian variant of MAML~\cite{finn2017model} using the Laplace approximation. 
Yoon \textit{et al.}~\cite{yoon2018bayesian} proposed a novel Bayesian MAML with a stein variational inference framework and chaser loss.
Finn \textit{et al.}~\cite{finn2018probabilistic} approximated MAP inference of the task-specific weights while maintain uncertainty only in the global weights.
Lee \textit{et al.}~\cite{lee2019learning} proposed a Bayesian meta-learning framework to deal with class/task imbalance and out-of-distribution tasks.
Lee \textit{et al.}~\cite{lee2019meta} proposed meta-dropout which generates learnable perturbations to regularize few-shot learning models.
In this paper, instead of modelling the uncertainty of tasks, we propose a novel Bayesian meta-learning framework to maximize the posterior distribution of domain augmentations.
\section{Method}

We first describe our problem setting and overall framework design. The goal is to learn a robust model from a {\it single} domain $\mathcal{S}$ and we expect the model to generalize over an {\it unknown} domain distribution $\{\mathcal{T}_1,\mathcal{T}_2,\cdots\} \sim p(\mathcal{T})$. This problem is more challenging than {\it domain adaptation} (assuming $p(\mathcal{T})$ is given) and {\it domain generalization} (assuming multiple source domains $\{\mathcal{S}_1,\mathcal{S}_2,\cdots\}$ are available). We create a series of domain augmentations $\{\mathcal{S}^+_1,\mathcal{S}^+_2,\cdots\} \sim p(\mathcal{S}^+)$ to approximate $p(\mathcal{T})$, from which the backbone $\theta$ can learn to generalize over unseen domains.

\begin{figure}
    \centering
    \includegraphics[width=.6\linewidth]{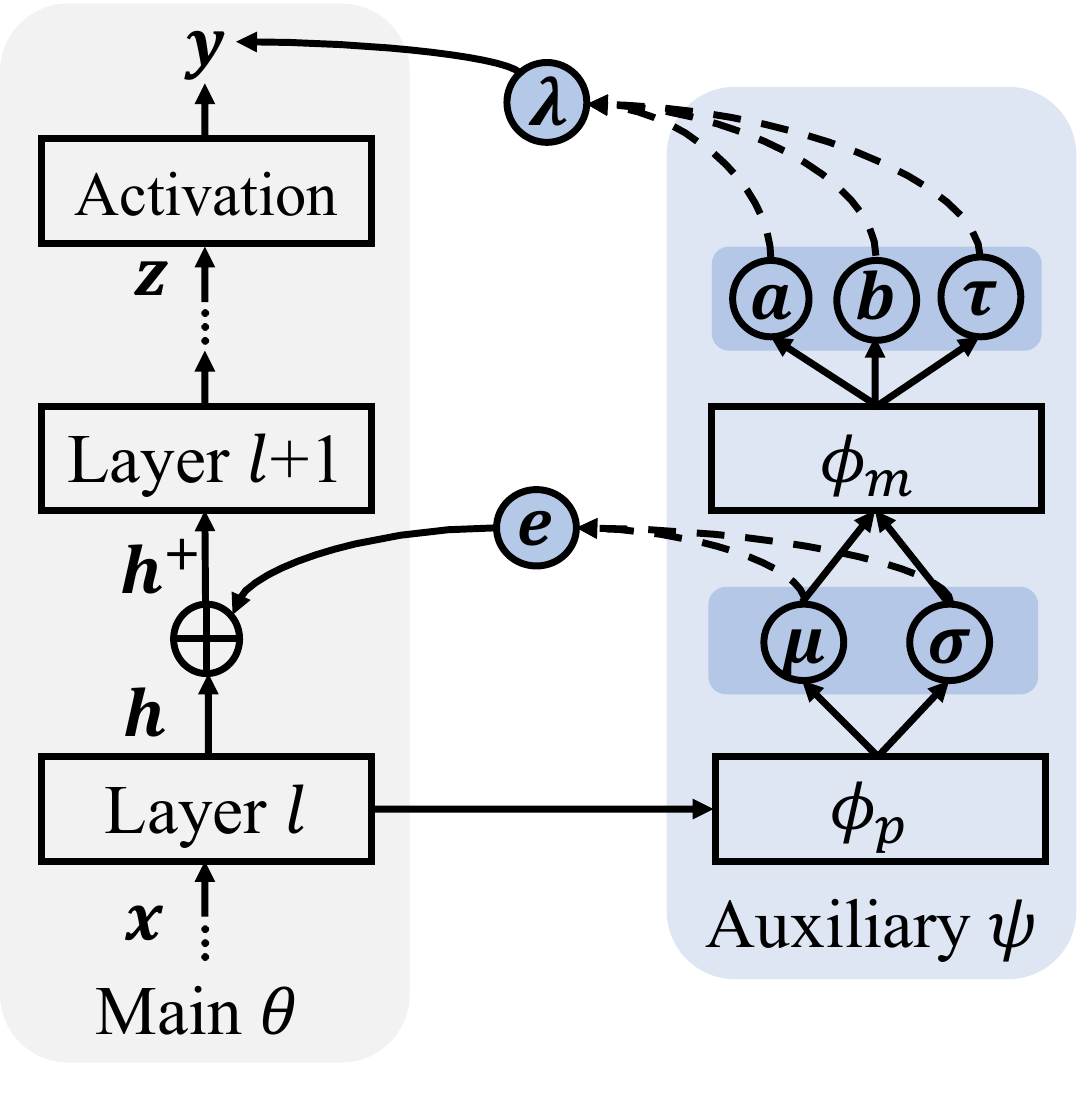}
    \caption{The main and auxiliary models.} \label{fig:models}
\end{figure}

{\bf Uncertainty-guided domain generalization.}
We assume that $\mathcal{S}^+$ should integrate uncertainty assessment for efficient domain generalization. To achieve it, we introduce the auxiliary $\psi=\{\phi_p,\phi_m\}$ to explicitly model the uncertainty with respect to $\theta$ and leverage it to create $\mathcal{S}^+$ by increasing the capacity in both input and label spaces. In input space, we introduce $\phi_p$ to create feature augmentations $\mathbf{h}^{+}$ via adding perturbation $\mathbf{e}$ sampled from $\mathcal{N}(\boldsymbol{\mu},\boldsymbol{\sigma})$. In label space, we integrate the same uncertainty encoded in  $(\boldsymbol{\mu},\boldsymbol{\sigma})$ into $\phi_m$ and propose learnable mixup to generate $\mathbf{y}^{+}$ (together with $\mathbf{h}^{+}$) through three variables $(a,b,\tau)$, yielding consistent augmentation in both input and output spaces.
To effectively organize domain augmentation and model training, we propose a Bayesian meta-learning framework to \textit{maximizing a posterior} of $p(\mathcal{S}^+)$ by jointly optimizing the backbone $\theta$ and the auxiliary $\psi$.
The overall framework is shown in Fig.~\ref{fig:models} and full algorithm is summarized in Alg.~\ref{alg:overrall}.

{\bf Merits of uncertainty assessment.}
Assessing the uncertainty of $\mathcal{S}^+$ plays a key role in our design. First, it provides consistent guidance to the augmentation in both input and label spaces when inferring $\mathcal{S}^+$,  which has never been studied before.
Second, we can gradually enlarge the domain transportation by increasing the uncertainty of $\mathcal{S}^+$ in a curriculum learning scheme~\cite{bengio2009curriculum}.
Last, we can easily assess the domain uncertainty by checking the value of $\boldsymbol{\sigma}$, which measures how unsure it is when deploying on unseen domains $\mathcal{T}$ (Sec.~\ref{sec:bayesian}).

\begin{algorithm}[t]
	\caption{Unseen Domain Generalization.}
	\LinesNumbered
	\label{alg:overrall}
	\KwIn{Source domain $\mathcal{S}$, \# of MC samples $K$.}
	\KwOut{Learned backbone $\theta$ and auxiliary $\psi$.}
	 \While{not converged}{
	     \textbf{Meta-train}: Compute $\theta^*$ on $\mathcal{S}$ using Eq.~\ref{eq:meta}
	     
	     Generate $\mathcal{S}^+$ from $\mathcal{S}$ using Eq.~\ref{eq:ada} 
	     
	     \For{$k=1,...,K$}{
	     Sample feature perturbation $\mathbf{h}^+_k$ using Eq.~\ref{eq:pertubation}
	     
	     Generate label mixup  $\mathbf{y}^+_k$  using Eq.~\ref{eq:mixup}
	     
	     \textbf{Meta-test}: Evaluate $\mathcal{L}(\theta^*;\mathcal{S}^+) $ w.r.t. $\mathcal{S}^+$
        }
        \textbf{Meta-update}: Update $\theta$ and $\psi$ using Eq.~\ref{eq:mc}
   }
\end{algorithm}

\subsection{Uncertainty-Guided Input Augmentation}\label{sec:input}


The goal is to create $\mathcal{S}^+$ from $\mathcal{S}$ such that $p(\mathcal{S^+})$ can approximate the out-of-domain distribution of $\mathcal{S}$. One the one hand, we expect a large domain transportation from $\mathcal{S}$ to $\mathcal{S}^+$ to best accommodate the unseen testing distribution $p(\mathcal{T})$. On the other hand, we prefer the transportation is domain-knowledge-free with uncertainty guarantee for broad and safe domain generalization. 
Towards this goal, we introduce $\phi_p$ to create feature augmentation $\mathbf{h}^+$  with large domain transportation through increasing the uncertainty with respect to $\theta$.

{\bf Adversarial Domain Augmentation.}
To encourage large domain transportation, we cast the problem in a worst-case scenario~\cite{sinha2017certifying} and propose to learn the auxiliary mapping $\phi_p$ via {\it adversarial domain augmentation}: 

\begin{equation}\label{eq:ada}
\underset{\phi_p}{\operatorname{maximize}} 
\underbrace{\mathcal{L}(\theta; \mathcal{S}^+)}_{\mathrm{Main \; task}} -  \beta\underbrace{\left\|\mathbf{z}-\mathbf{z}^+\right\|_{2}^{2}}_{\mathrm{Constraint}}.
\end{equation}

Here, $\mathcal{L}$ denotes empirical loss such as cross-entropy loss for classification.
The second term is the worst-case constraint, bounding the largest domain discrepancy between $\mathcal{S}$ and $\mathcal{S^+}$ in embedding space.
$\mathbf{z}$ denotes the FC-layer output right before the activation layer, which is distinguished from $\mathbf{h}$ that denotes the Conv-layer outputs. 

One merit of the proposed uncertainty-guided augmentation is that we can effectively relax the constraint to encourage large domain transportation in a curriculum learning scheme, which is significantly more efficient than ~\cite{qiao2020learning} that has to train an extra WAE-GAN ~\cite{tolstikhin2018wasserstein} to achieve this goal.
We introduce the detailed form of $\mathbf{h}^+$ as follows.

{\bf Variational feature perturbation.} 
To achieve adversarial domain augmentation, we apply uncertainty-guided perturbations to latent features instead of directly augmenting raw data, yielding domain-knowledge-free augmentation.
We propose to learn layer-wise feature perturbations $\mathbf{e}$ that transport latent features $\mathbf{h} \rightarrow \mathbf{h}^+$ for efficient domain augmentation $\mathcal{S} \rightarrow \mathcal{S}^+$. Instead of a direct generation $\mathbf{e}=f_{\phi_p}(\mathbf{x},\mathbf{h})$ widely used in previous work~\cite{volpi2018generalizing,qiao2020learning}, we assume $\mathbf{e}$ follows a multivariate Gaussian distribution $\mathcal{N}(\boldsymbol{\mu},\boldsymbol{\sigma})$, which can be used to easily access the uncertainty.
More specifically, the Gaussian parameters are learnable via variational inference $(\boldsymbol{\mu},\boldsymbol{\sigma})=f_{\phi_p}(\mathcal{S},\theta)$, such that:
\begin{equation}
\mathbf{h}^+ \leftarrow \mathbf{h} + \text{Softplus}(\mathbf{e})
\text{, where }
\mathbf{e} \sim \mathcal{N}(\boldsymbol{\mu},\boldsymbol{\sigma}),
\label{eq:pertubation}
\end{equation}
where $\text{Softplus}(\cdot)$ is applied to stabilize the training.
$\phi_p$ can create a series of feature augmentations $\{\mathbf{h}^+_1, \mathbf{h}^+_2,\cdots\}$ in different training iterations.  In Sec.~\ref{sec:ablation}, we empirically show that $\{\mathbf{h}^+_1, \mathbf{h}^+_2,\cdots\}$ gradually enlarge the transportation through increasing the uncertainty of augmentations in a curriculum learning scheme and enable the model to learn from ``easy'' to ``hard'' domains.

\subsection{Uncertainty-Guided Label Mixup}\label{sec:mix}

Feature perturbations not only augment the input but also yield label uncertainty. 
To explicitly model the label uncertainty, we leverage the input uncertainty, encoded in  $(\boldsymbol{\mu},\boldsymbol{\sigma})$, 
to infer the label uncertainty encoded in $(a,b,\tau)$ through $\phi_m$ as shown in Fig.~\ref{fig:models}.
We leverage the label uncertainty to propose learnable label mixup, yielding consistent augmentation in both input and output spaces and further reinforcing generalization over unseen domains.

{\bf Random Mixup.} We start by introducing random {\it mixup}~\cite{zhang2017mixup} for robust learning. The key idea is to regularize the training to favor simple linear behavior in-between examples. More specifically, {\it mixup} performs training on convex interpolations of pairs of examples ($\mathbf{x}_i,\mathbf{x}_j$) and their labels ($\mathbf{y}_i,\mathbf{y}_j$):
\begin{equation*}
\mathbf{x}^+=\lambda \mathbf{x}_{i}+(1-\lambda) \mathbf{x}_{j}, \quad \mathbf{y}^+=\lambda \mathbf{y}_{i}+(1-\lambda) \mathbf{y}_{j}, 
\end{equation*}
where $\lambda \sim \operatorname{Beta}(\alpha,\alpha)$ and the {\it mixup} hyper-parameter $\alpha \in (0, +\infty)$ controls the interpolation strength.

{\bf Learnable Label Mixup.} 
We improve {\it mixup} by casting it in a learnable framework specially tailored for single source generalization.
First, instead of mixing up pairs of examples, we mix up $\mathcal{S}$ and $\mathcal{S}^+$ to achieve in-between domain interpolations. Second, we leverage the uncertainty encoded in $(\boldsymbol{\mu},\boldsymbol{\sigma})$ to predict learnable parameters $(a,b)$, which controls the direction and strength of domain interpolations: 
\begin{equation}\label{eq:mixup}
\mathbf{h}^+ =\lambda \mathbf{h} + (1-\lambda) \mathbf{h}^+, \quad \mathbf{y}^+ =\lambda \mathbf{y}+(1-\lambda) \tilde{\mathbf{y}},
\end{equation}
where $\lambda \sim \operatorname{Beta}(a, b)$ and $\tilde{\mathbf{y}}$ denotes a {\it label-smoothing}~\cite{szegedy2016rethinking} version of $\mathbf{y}$. More specifically, we perform {\it label smoothing} by a chance of $\tau$, such that we assign $\rho \in(0,1)$ to the true category and equally distribute $\frac{1-\rho}{c-1}$ to the others, where $c$ counts categories. The Beta distribution $(a,b)$ and the lottery $\tau$ are jointly inferred by $(a,b,\tau)=f_{\phi_m}(\boldsymbol{\mu},\boldsymbol{\sigma})$ to integrate the uncertainty of domain augmentation.

\subsection{A Unified Framework}\label{sec:bayesian}

To effectively organize domain augmentation and model training, we propose a Bayesian meta-learning framework to \textit{maximize a posterior} of $p(\mathcal{S}^+)$ by jointly optimizing the backbone $\theta$ and the auxiliary $\psi=\{\phi_p,\phi_m\}$.
Specifically, we {\it meta-train} the backbone $\theta$ on the source $\mathcal{S}$ and {\it meta-test} its generalization capability over $p(\mathcal{S}^+)$, where $\mathcal{S}^+$ is 
generated by performing data augmentation
in both input (Sec.~\ref{sec:input}) and output (Sec.~\ref{sec:mix}) spaces through the auxiliary $\psi$. 
Finally, we {\it meta-update} $\{\theta,\psi\}$ using gradient:
\begin{equation}
\nabla_{\theta,\psi} \mathbb{E}_{p(\mathcal{S}^+)} [\mathcal{L}( \theta^*;\mathcal{S^+})] \text{,where } \theta^* \equiv \theta - \alpha \nabla_{\theta}\mathcal{L}(\theta;\mathcal{S}).
\label{eq:meta}
\end{equation}
Here $\theta^*$ is the meta-trained backbone on $\mathcal{S}$ and $\alpha$ is the learning rate.
After training, the backbone $\theta$ is expected to bound the generalization uncertainty over unseen populations  $p(\mathcal{T})$ in a worst-case scenario (Sec.~\ref{sec:input}) while $\psi$ can be used to access the value of uncertainty efficiently. 

{\bf Bayesian Meta-learning.}
The goal is to maximize the conditional likelihood of the augmented domain $\mathcal{S^+}$: $\log p\left(\mathbf{y^+}| \mathbf{x}, \mathbf{h}^+; \theta^*\right)$. However, solving it involves the true posterior $p\left(\mathbf{h}^+ | \mathbf{x} ; \theta^*,\psi\right)$, which is intractable~\cite{lee2019learning}. Thus, we resort to amortized variational inference with
a tractable form of approximate posterior $q\left(\mathbf{h}^+ | \mathbf{x} ; \theta^*,\psi\right)$. 
The approximated lower bound is as follows:
\begin{equation}
L_{\theta, \psi}= \mathbb{E}_{q\left(\mathbf{h}^+ | \mathbf{x} ; \theta^*,\psi\right)}[\log \frac{p\left(\mathbf{y^+}| \mathbf{x}, \mathbf{h}^+; \theta^*\right)}{q\left(\mathbf{h}^+ | \mathbf{x} ; \theta^*,\psi\right)}].
\label{eq:lower}
\end{equation}

We leverage Monte-Carlo (MC) sampling to maximize the lower bound $L_{\theta, \psi}$ by:
\begin{equation}
\begin{split}
\min _{\theta, \psi} &\frac{1}{K} \sum_{k=1}^{K} \left[-\log p\left(\mathbf{y}^+_k | \mathbf{x}, \mathbf{h}^+_k; \theta^*\right)\right]+\\
&\mathrm{KL}\left[q\left(\mathbf{h}^+ | \mathbf{x} ; \theta^*,\psi\right) \| p\left(\mathbf{h}^+ | \mathbf{x} ; \theta^*,\psi\right)\right],
\label{eq:mc}
\end{split}
\end{equation}
where $\mathbf{h}^+_k \sim q\left(\mathbf{h}^+ | \mathbf{x} ; \theta^*,\psi\right)$ and $K$ is the number of MC samples. 
Instead of setting the prior distribution to $\mathcal{N}(\mathbf{0},\mathbf{I})$ in~\cite{kingma2013auto},
we assume that $q\left(\mathbf{h}^+ | \mathbf{x} ; \theta^*,\psi\right)$ is expected to approximate $p\left(\mathbf{h}^+ | \mathbf{x} ; \theta^*,\psi\right)$ through the adversarial training on $\phi_p$ in Eq.~\ref{eq:ada}. 
Thank to the Bayesian meta-learning framework, the generalization uncertainty on unseen domains is significantly suppressed (Sec.~\ref{sec:ablation}). More importantly, a few examples of the target domain can quickly adapt $\theta$ to be domain-specific, yielding largely improved performance for few-shot domain adaptation (Sec.~\ref{sec:image}).   

{\bf Uncertainty Estimation.} At testing time, given a novel domain $\mathcal{T}$, we propose a {\it normalized domain uncertainty score}, $| \frac{\boldsymbol{\sigma}(\mathcal{T}) - \boldsymbol{\sigma}(\mathcal{S})}{\boldsymbol{\sigma}(\mathcal{S})}|$, to estimate its uncertainty with respect to learned $\theta$. Considering $\psi$ is usually much smaller than $\theta$, this score can be calculated efficiently by one-pass data forwarding through $\psi$. In Sec.~\ref{sec:image}, we empirically prove that our estimation is consistent with conventional Bayesian methods~\cite{blundell2015weight}, while the time consumption is significantly reduced by an order of magnitude.

\begin{table*}[t!]
	\begin{minipage}[t!]{1.0\textwidth}
	\begin{center}
	    \label{tab:digits}
		\resizebox{0.9\linewidth}{!}{
		\begin{tabular}{@{}l|cccc|ccc|cccc@{}}
			\toprule
			\toprule
			Domain  & Mixup~\cite{zhang2017mixup} & PAR~\cite{wang2019learning}  & Self-super~\cite{hendrycks2019using} &  JiGen  \cite{carlucci2019jigasaw} & ERM~\cite{koltchinskii2011oracle}  & GUD~\cite{volpi2018generalizing} & M-ADA~\cite{qiao2020learning} & \textbf{Ours} \\
			\midrule
			 SVHN~\cite{lecun1998gradient}  & 28.5 & 30.5 & 30.0 & 33.8 & 27.8 & 35.5 & \underline{42.6} & \textbf{43.3} \\
			 MNIST-M~\cite{ganin2015unsupervised}  & 54.0  & 58.4 & 58.1 & 57.8 & 52.8 & 60.4 & \textbf{67.9 } & \underline{67.4} \\
			 SYN~\cite{ganin2015unsupervised}  & 41.2 & 44.1 & 41.9 & 43.8 & 39.9 & 45.3 & \underline{49.0} & \textbf{57.1} \\
			 USPS~\cite{denker1989advances}  & 76.6 & 76.9 & 77.1 & 77.2 & 76.5 & 77.3 & \textbf{78.5}  & \underline{77.4} \\
			 \midrule
			 Avg.  & 50.1& 52.5 & 51.8 & 53.1 & 49.3 & 54.6 & \underline{59.5} & \textbf{61.3} \\
			\bottomrule
	\end{tabular}}
	\end{center}
	\end{minipage}
	\begin{minipage}[t!]{1.0\textwidth}
	    \vspace{0.1in}
	    \begin{center}
	    \label{tab:cifar}
	\resizebox{0.9\linewidth}{!}{
		\begin{tabular}{@{}l|cccc|ccc|cccc@{}}
			\toprule
			Model   & Mixup~\cite{zhang2017mixup} & Cutout~\cite{devries2017cutout}  & AutoAug~\cite{cubuk2019autoaugment} & PGD~\cite{madry2017pgd} & ERM~\cite{koltchinskii2011oracle} & GUD~\cite{volpi2018generalizing} & M-ADA~\cite{qiao2020learning} & \textbf{Ours} \\
			\hline
			 AllConv~\cite{salimans2016weight}  & 75.4 & 67.1  & 70.8 & 71.9 & 69.2 & 73.6 & \underline{75.9} & \textbf{79.6} \\
			 WRN~\cite{zagoruyko2016wide}  & 77.7 & 73.2  & 76.1 & 73.8 & 73.1 & 75.3 & \underline{80.2}  & \textbf{83.4} \\
			\bottomrule
	\end{tabular}}
	\end{center}
	\end{minipage}
	\captionof{table}{Image classification accuracy (\%) on {\it Digits}~\cite{volpi2018generalizing} (\textbf{top}) and {\it CIFAR-10-C}~\cite{hendrycks2019benchmarking} (\textbf{bottom}). We compare with {\it robust training} (\textbf{Columns 1-4}) and {\it domain generalization} (\textbf{Columns 5-7}). For {\it Digits}, all models are trained on {\it MNIST}~\cite{lecun1998gradient}. For {\it CIFAR-10-C}, two widely employed backbones are evaluated. Our method outperforms M-ADA~\cite{qiao2020learning} (previous SOTA) consistently in all settings.} \label{tab:image}
    \vspace{0.1in}
\end{table*}

\section{Experiments}
\begin{figure*}
\centering
\begin{minipage}[t!]{0.60\textwidth}
\centering
\vspace{-0.1in}
\subfigure{\includegraphics[width=0.49\linewidth]{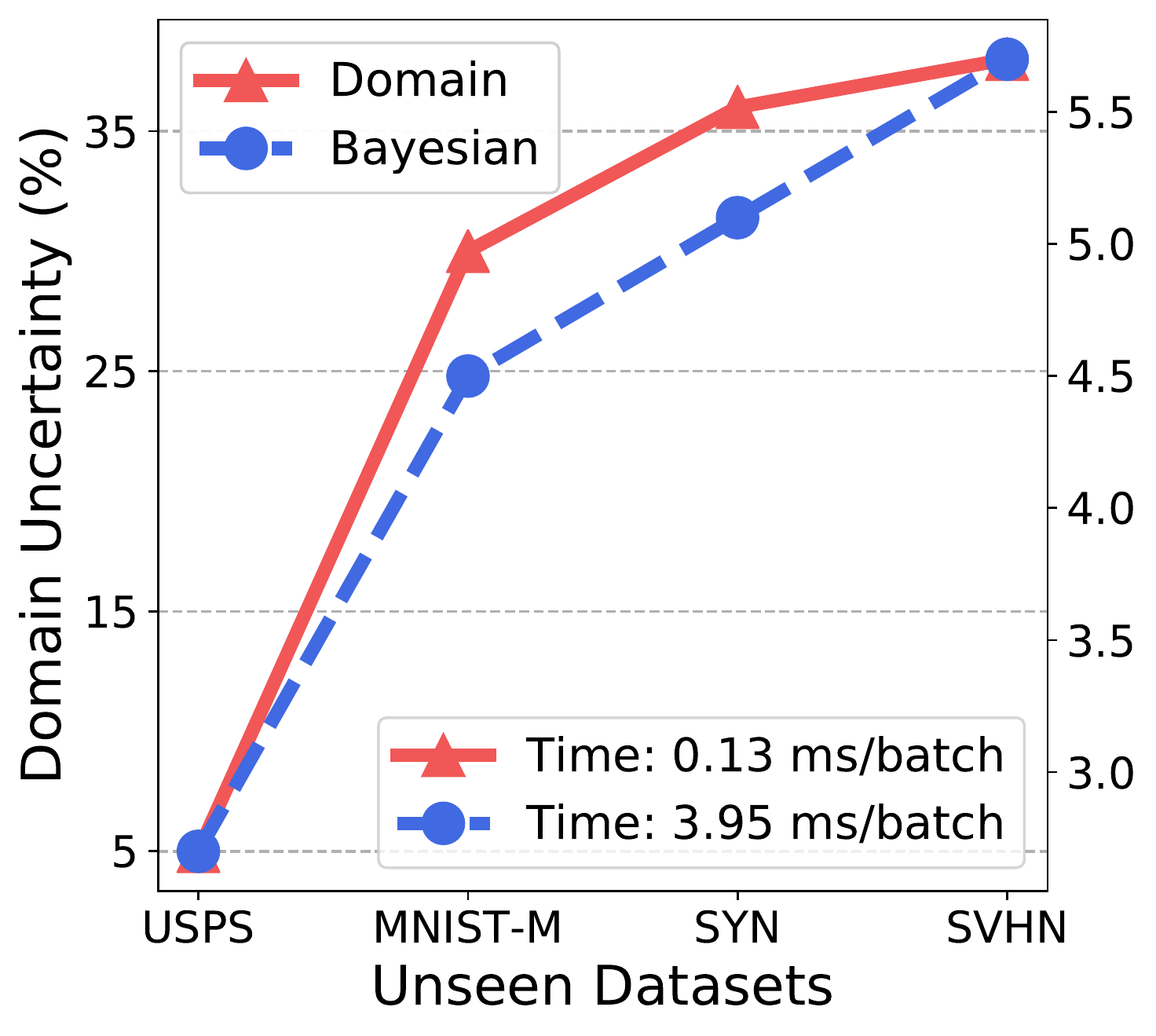}}
\hspace{-0.1in}
\subfigure{\includegraphics[width=0.49\linewidth]{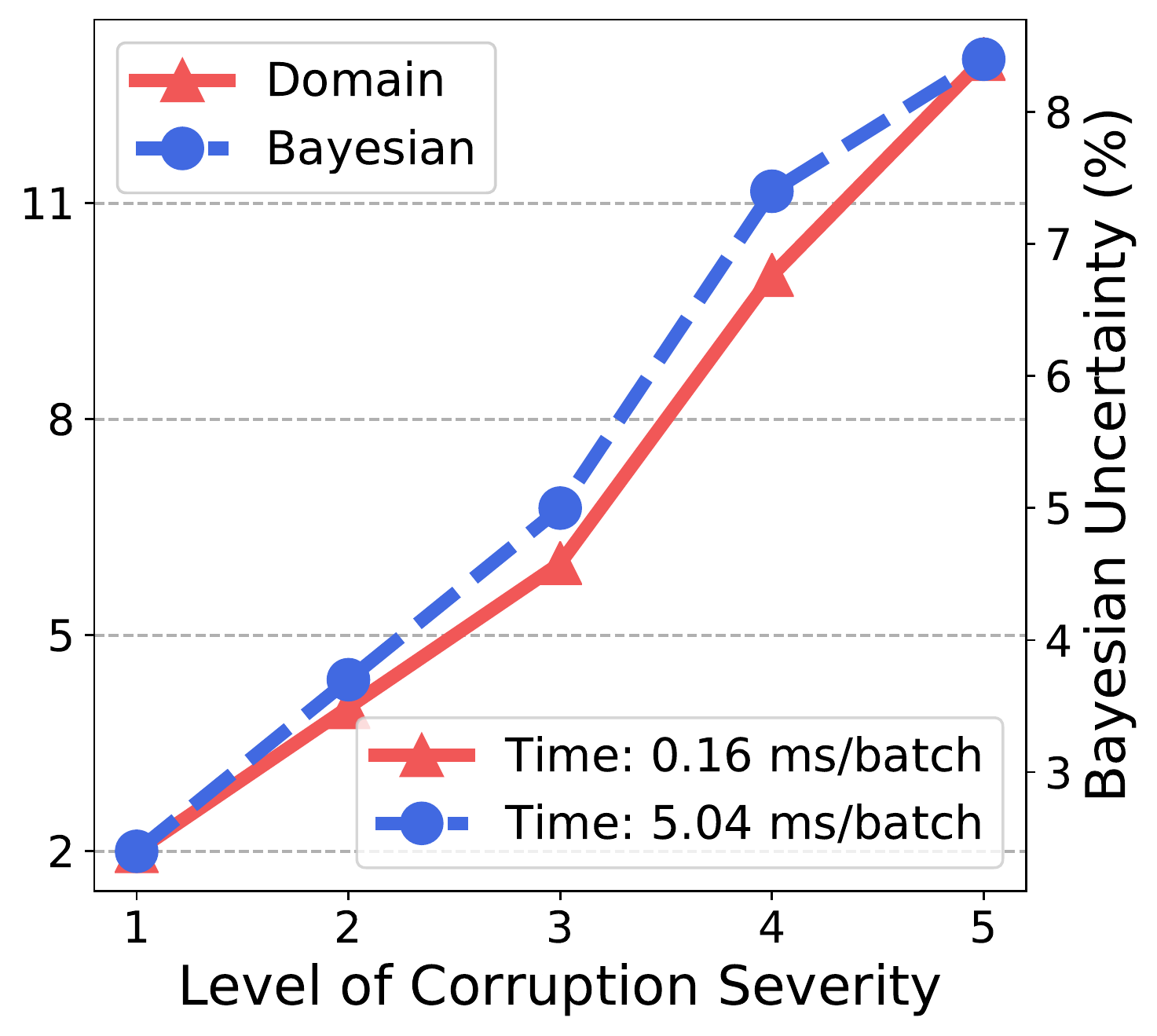}}
\vspace{-0.2in}
\caption{Uncertainty estimation on {\it Digits} (\textbf{left}) and {\it CIFAR-10-C} (\textbf{right}).
Our prediction of \textit{domain uncertainty} is consistent with \textit{Bayesian uncertainty}, while our method is an order of magnitude faster since we forward data only once.}
\label{fig:uncertainty}
\end{minipage}
\hspace{0.1in}
\begin{minipage}{0.32\textwidth}
\centering
\resizebox{\linewidth}{!}{
		\begin{tabular}[]{@{}l|c|cc@{}}
			\toprule
			Method & $\lvert  \mathcal{T} \rvert$ & M \ $\rightarrow$ S & Avg.\\
			\hline
			DIRT-T~\cite{shu2018dirt} & \multirow{3}{*}{All}&  54.5 &-\\
			SE~\cite{french2017self} & &14.0 & 70.4\\
			SBADA~\cite{russo2018source} & & \textbf{61.1} & 78.3\\
			\hline
			FADA~\cite{motiian2017few} & 7  & 47.0 &   75.2\\
			CCSA~\cite{motiian2017unified} & 10 &37.6 &  76.0\\
			\hline
			\multirow{2}{*}{\bf Ours} &7  & 58.1  & \underline{80.1}\\
			&10 & \underline{59.8} & \textbf{81.5} \\
			\bottomrule
\end{tabular}}
\captionof{table}{Few-shot domain adaptation accuracy (\%) on {\it MNIST(M), USPS(U), and SVHN(S)}. $\lvert \mathcal{T} \rvert$ denotes the number of target samples (per class) used during model training.}\label{tab:few_da}
\end{minipage}
\end{figure*}

To best validate the performance, we conduct a series of experiments to compare our approach with existing methods that can be roughly grouped in four categories: {\bf 1) Adversarial training:} PAR~\cite{wang2019learning}, Self-super~\cite{hendrycks2019using}, and PGD~\cite{madry2017pgd}. {\bf 2) Data augmentation:} Mixup~\cite{zhang2017mixup}, JiGen~\cite{carlucci2019jigasaw}, Cutout~\cite{devries2017cutout}, and AutoAug~\cite{cubuk2019autoaugment}. {\bf 3) Domain adaptation:} DIRT-T~\cite{shu2018dirt}, SE~\cite{french2017self}, SBADA~\cite{russo2018source}, FADA~\cite{motiian2017few}, and CCSA~\cite{motiian2017unified}. {\bf 4) Domain generalization:} ERM~\cite{koltchinskii2011oracle}, GUD~\cite{volpi2018generalizing}, and M-ADA~\cite{qiao2020learning}.
The experimental results prove that our method achieves superior performance on a wide scope of tasks, including {\it image classification}~\cite{hendrycks2019benchmarking}, {\it semantic segmentation}~\cite{ros2016synthia}, {\it text classification}~\cite{chen2012marginalized}, and {\it speech recognition}~\cite{warden2018speech}. Please refer to supplementary for more details about experiment setup.

\subsection{Image Classification}\label{sec:image}

\begin{table*}[th]
\begin{center}
\resizebox{.9\linewidth}{!}{
\begin{tabular}{@{}llccccccccccc@{}}
\toprule
& &\multicolumn{5}{c}{New York-like City}& \multicolumn{5}{c}{Old European Town}& \\
\cmidrule(lr){3-7} \cmidrule(lr){8-12}
Source Domain& Method &Dawn&Fog&Night&Spring&Winter& Dawn&Fog&Night&Spring&Winter& Avg. \\
\hline
\multirow{3}*{Highway/Dawn}&ERM \cite{koltchinskii2011oracle}&27.8 &2.7 &0.9 &6.8 &1.7 &52.8 &31.4 &15.9 &33.8 &13.4 &18.7 \\
&GUD~\cite{volpi2018generalizing}&27.1 &4.1 &1.6 &7.2 &2.8 & 52.8 &34.4 &18.2 &33.6 &14.7 &19.7 \\
&M-ADA~\cite{qiao2020learning}& \underline{29.1} & \underline{4.4}& \textbf{4.8}& \textbf{14.1} &\underline{5.0}&\underline{54.3}&\underline{36.0}&\underline{23.2}&\textbf{37.5}&\underline{14.9}&\underline{22.3} \\
&{\bf Ours}& \textbf{29.3} & \textbf{7.6} & \underline{2.8} & \underline{12.7} & \textbf{10.2} & \textbf{54.9} & \textbf{37.0} & \textbf{25.3} & \underline{37.2} & \textbf{17.7} & \textbf{23.5}  \\
\hline
\multirow{3}*{Highway/Fog}&ERM \cite{koltchinskii2011oracle}& 17.2&34.8&12.4&26.4&11.8&33.7&55.0&26.2&41.7&12.3&27.2 \\
&GUD \cite{volpi2018generalizing}& 18.8&\underline{35.6}&\underline{12.8}&26.0&13.1&37.3&\underline{56.7}&28.1&\underline{43.6}&\textbf{13.6}&28.5 \\
&M-ADA~\cite{qiao2020learning}& \underline{21.7}&32.0&9.7&\underline{26.4}&\underline{13.3}&\underline{42.8}&56.6&\textbf{31.8}&42.8&12.9&\underline{29.0} \\
&{\bf Ours}&\textbf{23.0} &\textbf{36.2}  &\textbf{13.5}  &\textbf{27.6} &\textbf{14.2} &\textbf{43.1} &\textbf{57.4} &\underline{31.0} &\textbf{44.6} &\underline{13.1} &\textbf{ 30.4 } \\
\bottomrule
\end{tabular}}
\end{center}
\caption{Semantic segmentation mIoU (\%) on \textit{SYNTHIA}~\cite{ros2016synthia}. All models are trained on the single source from \textit{Highway} and evaluated on unseen environments from \textit{New York-like City} and \textit{Old European Town}.}\label{tab:seg}
\end{table*}

{\bf Datasets.}
We validate our method on the following two benchmark datasets for image classification. 
(1) {\it Digits} is used for digit classification and 
consists of five sub-datasets: MNIST~\cite{lecun1998gradient}, MNIST-M~\cite{ganin2015unsupervised}, SVHN~\cite{netzer2011reading}, SYN~\cite{ganin2015unsupervised}, and USPS~\cite{denker1989advances}. Each sub-dataset can be viewed as a different domain. 
Each image in these datasets contains one single digit with different styles and backgrounds.  
(2) {\it CIFAR-10-C}~\cite{hendrycks2019benchmarking} is a robustness benchmark consisting of 19 corruptions types with five levels of severity applied to the test set of CIFAR-10~\cite{krizhevsky2009learning}.
The corruptions consist of four main categories: noise, blur, weather, and digital. Each corruption has five-level severities and ``5'' indicates the most corrupted one.

{\bf Setup.} {\it Digits}: following the setup in \cite{volpi2018generalizing},
we use 10,000 samples in the training set of MNIST for training, and evaluate models on the other four sub-datasets. We use a ConvNet \cite{lecun1989backpropagation} with architecture \textit{conv-pool-conv-pool-fc-fc-softmax} as the backbone. 
All images are resized to 32$\times$32, and the channels of MNIST and USPS are duplicated to make them as RGB images.
{\it CIFAR-10-C}: we train models on CIFAR-10 and evaluate them on CIFAR-10-C. Following the setting of~\cite{hendrycks2019augmix}, we evaluate the model on 15 corruptions. We train models on AllConvNet (AllConv)~\cite{salimans2016weight} and Wide Residual Network (WRN)~\cite{zagoruyko2016wide} with 40 layers and width of 2.

\begin{table*}[th]
	\begin{minipage}[]{0.48\textwidth}
        \begin{center}
		\resizebox{1.0\linewidth}{!}{
		\begin{tabular}{@{}lcccccc@{}}
			\toprule
			       & \multicolumn{3}{c}{books} & \multicolumn{3}{c}{dvd}   \\
			\cmidrule(lr){2-4} \cmidrule(lr){5-7}
			Method    & d & k & e & b & k & e  \\
			\hline
			ERM~\cite{koltchinskii2011oracle}    & 78.7  & 74.6 & 63.6 & 78.5 & 82.1 & \textbf{75.2}  \\
			GUD~\cite{volpi2018generalizing}    & 79.1 & 75.6 &  64.7 & 78.1 & 82.0 & 74.6\\
			M-ADA~\cite{qiao2020learning}    & \underline{79.4} & \underline{76.1} & \underline{65.3}  & \underline{78.8} & \underline{82.6}  & 74.3 \\
			\hline
		    \textbf{Ours}    & \textbf{80.2}  & \textbf{76.8} &\textbf{67.1} & \textbf{80.1} & \textbf{83.5} & \underline{75.0} \\
			\bottomrule
	    \end{tabular}}
	    \captionof{table}{Text classification accuracy (\%) on {\it Amazon Reviews}. Models are trained on one text domain and evaluated on unseen text domains. Our method outperforms others in all settings except ``$dvd \rightarrow electronics$''.}\label{tab:text}
	    \end{center}
	 \end{minipage}
	\hspace{0.1in}
	\begin{minipage}[]{0.48\textwidth}
	    \begin{center}
	    \label{tab:speech}
	    \resizebox{1.0\linewidth}{!}{
		\begin{tabular}{@{}lcccccc@{}}
			\toprule
			       & \multicolumn{2}{c}{Time} & \multicolumn{3}{c}{Frequency}   \\
			\cmidrule(lr){2-3} \cmidrule(lr){4-6}
			Method    & Amp. & Pit. & Noise & Stretch & Shift   \\
			\hline
			ERM~\cite{koltchinskii2011oracle}     & 63.8  & 71.6 & 73.9 & 72.9 & 70.5   \\
			GUD~\cite{volpi2018generalizing}  & 64.1 & \underline{72.1} & 74.8  & 73.1  &  70.9 \\
			M-ADA~\cite{qiao2020learning} & \underline{64.5} & 71.9 &  \underline{75.4} & \underline{73.8} & \underline{71.4}    \\
			\hline
		    \textbf{Ours}    & \textbf{65.3}  & \textbf{73.5} &\textbf{75.8}   & \textbf{75.0} & \textbf{72.5} \\
			\bottomrule
	    \end{tabular}}
	     \captionof{table}{Speech recognition accuracy (\%) on {\it Google Commands}. Models are trained on clean set and evaluated on five corrupted sets. Results validate our strong generalization on corruptions in both time and frequency domains.}\label{tab:speech}
	    \end{center}
    \end{minipage}
\end{table*}

{\bf Results.} {\bf 1) Classification accuracy.} Tab.~\ref{tab:image} shows the classification results of {\it Digits} and {\it CIFAR-10-C}. On the experiment of {\it Digits}, GUD~\cite{volpi2018generalizing}, M-ADA~\cite{qiao2020learning}, and our method outperform all baselines of the second block. And our method outperforms M-ADA~\cite{qiao2020learning} on {\it SYN} and the average accuracy by 8.1\% and 1.8\%, respectively. On the experiment of {\it CIFAR-10-C}, our method consistently outperforms all baselines on two different backbones, suggesting its strong generalization on various image corruptions.
{\bf 2) Uncertainty estimation.}
We compare the proposed \textit{domain uncertainty score} (Sec.\ref{sec:bayesian}) with a more time-consuming one based on Bayesian models~\cite{blundell2015weight}. The former computes the uncertainty through one-pass forwarding, while the latter computes the variance of the output through repeated sampling of 30 times. Fig.~\ref{fig:uncertainty} show the results of uncertainty estimation on {\it Digits} and  {\it CIFAR-10-C}. As seen, our estimation shows consistent results with Bayesian uncertainty estimation on both {\it Digits} and  {\it CIFAR-10-C}, suggesting its high efficiency. 
{\bf 3) Few-shot domain adaptation.} Although our method is designed for single domain generalization, we also show that our method can be easily applied for few-shot domain adaptation~\cite{motiian2017few} due to the meta-learning training scheme. Following the setup in~\cite{qiao2020learning}, the model is first pre-trained on the source domain $\mathcal{S}$ and then fine-tuned on the target domain $\mathcal{T}$.
We conduct three few-shot domain adaption tasks: \textit{USPS(U)}$\rightarrow$\textit{MNIST(M)}, \textit{MNIST(M)}$\rightarrow$\textit{SVHN(S)}, and \textit{SVHN(S)}$\rightarrow$\textit{MNIST(M)}.
Results of the three tasks are shown in Tab.~\ref{tab:few_da}. 
Our method achieves the best performance on the average of three tasks. 
The result on the hardest task (\textit{M}$\rightarrow$\textit{S}) is even competitive to that of SBADA~\cite{russo2018source} which uses all images of the target domain for training. Full results are provided in supplementary.

\subsection{Semantic Segmentation}\label{sec:seg}

{\bf Datasets.}
{\it SYTHIA}~\cite{ros2016synthia} is a synthetic dataset of urban scenes, used for semantic segmentation in the context of driving scenarios. 
This dataset consists of photo-realistic frames rendered from virtual cities and comes with precise pixel-level semantic annotations.
It is composed of the same traffic situation but under different locations (Highway, New York-like City, and Old European Town are selected) and different weather/illumination/season conditions (Dawn, Fog, Night, Spring, and Winter are selected).

{\bf Setup.} 
In this experiment, Highway is the source domain, and New York-like City together with Old European Town are unseen domains. Following the protocol in \cite{volpi2018generalizing,qiao2020learning}, we only use the images from the left front camera and 900 images are randomly sample from each source domain. We use FCN-32s~\cite{long2015fully} with the backbone of ResNet-50~\cite{he2016deep}.

{\bf Results.}
We report the mean Intersection Over Union (mIoU) of {\it SYTHIA} in Tab.~\ref{tab:seg}. As can be observed, our method outperforms previous SOTA in most unseen environments.
Results demonstrate that our model can better generalize to the changes of locations, weather, and time.
We provide visual comparison in the supplementary.

\subsection{Text Classification}


{\bf Datasets.}
{\it Amazon Reviews}~\cite{chen2012marginalized} contains reviews of products belonging to four categories - books(b), DVD(d), electronics(e) and kitchen appliances(k). The difference in textual description of the four product categories manifests as domain shift. Following~\cite{ganin2015unsupervised}, we use unigrams and bigrams as features resulting in 5000 dimensional representations.

{\bf Setup.} We train the models on one source domain (books or dvd), and evaluate them on the other three domains.
Similar to~\cite{ganin2015unsupervised}, we use a neural network with two hidden layers (both with 50 neurons) as the backbone. 

{\bf Results.}
Tab.~\ref{tab:text} shows the results of text classification on {\it Amazon Reviews}~\cite{chen2012marginalized}. It appears that our method outperform previous ones on all the three unseen domains when the source domain is ``books''. We note that there is a little drop in performance on ``electronics'' when the source domain is ``dvd''. One possible reason is that ``electronics'' and  ``dvd'' may share a similar distribution. And our method creates large distribution shift, degrading the performance on ``electronics''.

\begin{figure*}
\centering
\begin{minipage}[h]{0.65\textwidth}
\centering
\includegraphics[width=1.\linewidth]{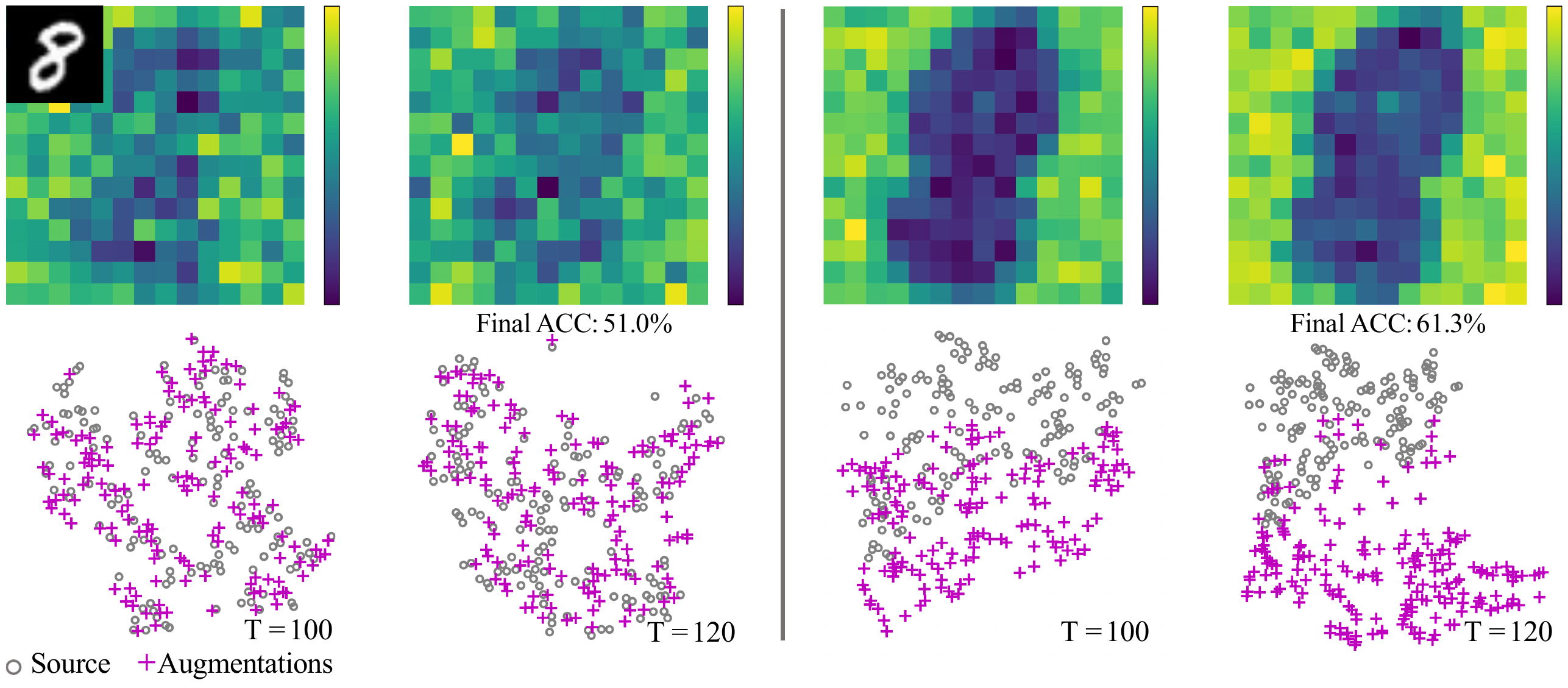}
\caption{Visualization of feature perturbation $|\mathbf{e}| = |\mathbf{h}^+ -\mathbf{h}|$ (\textbf{Top}) and embedding of domains (\textbf{Bottom}) at different training iterations $T$ on \textit{MNIST}.
\textbf{Left:} Models w/o uncertainty; \textbf{Right:} Models w/ uncertainty. Most perturbations are located in the background area and models w/ uncertainty can create large domain transportation in a curriculum learning scheme.}\label{fig:feature}
\end{minipage}
\hspace{0.1in}
\begin{minipage}{0.32\textwidth}
\centering
\includegraphics[width=1.\linewidth]{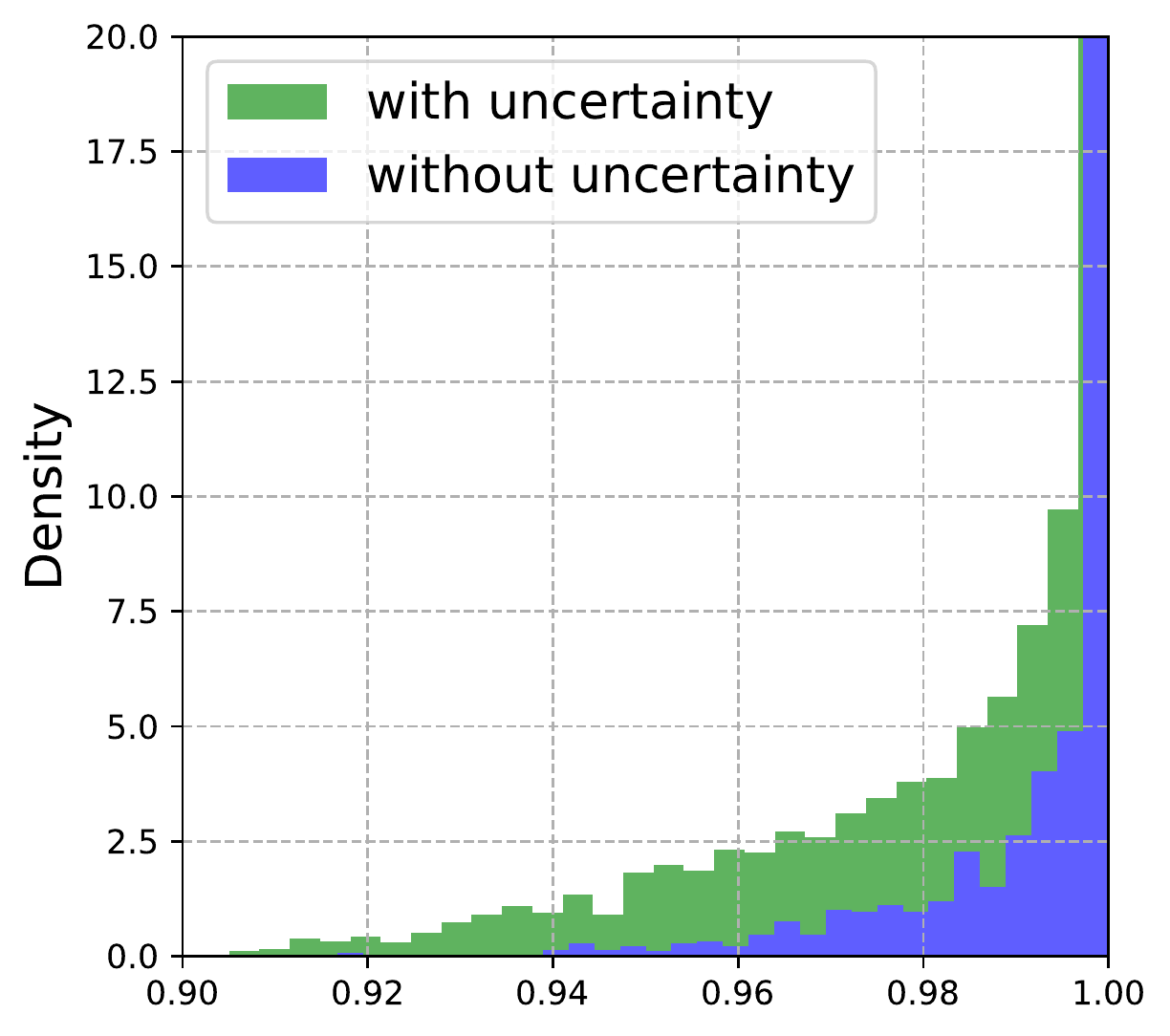}
\caption{Visualization of label mixup $\mathbf{y}^+$ on \textit{MNIST}. Models w/ uncertainty can encourage more smoothing labels and significantly increase the capacity of label space.}\label{fig:y_plus}
\end{minipage}
\end{figure*}

\subsection{Speech Recognition}


{\bf Datasets.} {\it Google Commands}~\cite{warden2018speech} contains 65000 utterances (one second long) from thousands of people. The goal is to classify them to 30 command words. There are 56196, 7477, and 6835 examples for training, validation, and test. To simulate domain shift in real-world scenario, we apply five common corruptions in both time and frequency domains. This creates five test sets that are ``harder'' than training sets, namely amplitude change (Amp.), pitch change (Pit.), background noise (Noise), stretch (Stretch), and time shift (Shift). 
In detail, the range of ``amplitude change'' is (0.7,1.1). The maximum scales of ``pitch change'', ``background noise'', and ``stretch'' are 0.2, 0.45, and 0.2, respectively. The maximum shift of ``time shift'' is 8.

{\bf Setup.} We train the models on the clean train set, and evaluate them on the corrupted test sets. We encode each audio into a mel-spectrogram with the size of 1x32x32 and feed them to LeNet~\cite{Lecun98gradient} as one-channel input. 

{\bf Results.}
Tab.~\ref{tab:speech} shows the results of speech recognition on {\it Google Commands}~\cite{warden2018speech}. Our method outperforms the other three methods on all the five corrupted test sets, indicating its strong generalization ability in both time and frequency domain. 
In detail, our method outperforms the second best by 0.8\% on ``amplitude change'', 1.4\% on ``pitch change'', 0.4\% on ``background noise'', 1.2\% on ``stretch'', and 1.1\% on ``time shift'', respectively. We can see that the improvements on ``pitch change'', ``stretch'', and ``time shift'' are more significant than those on ``amplitude change'' and ``background noise''.

\subsection{Ablation Study}\label{sec:ablation}

\begin{table}[]
    \centering
   	\resizebox{.9\linewidth}{!}{
		\begin{tabular}{@{}lcc@{}}
			\toprule
			   & Digits~\cite{volpi2018generalizing} & CIFAR-10-C~\cite{hendrycks2019benchmarking} \\
			\hline
			\textbf{Full Model} & \textbf{61.3}$\pm$0.73 &  \textbf{70.2}$\pm$0.62\\
			\hline
			Random Gaussian & 51.0$\pm$0.36  & 64.0$\pm$0.18 \\
			Determ. perturb. & 59.7$\pm$0.70 & 67.0$\pm$0.57 \\
			Random $\boldsymbol{\mu}$ & 60.5$\pm$0.75 & 69.1$\pm$0.61 \\
			Random $\boldsymbol{\sigma}$ & 60.7$\pm$0.65 & 69.5$\pm$0.60 \\
			\bottomrule
	    \end{tabular}}
    \caption{Ablation study of feature perturbation.}\label{tab:ablation_x}
\end{table}

In this section, we perform ablation study to investigate key components of our method. For {\it Digits}~\cite{volpi2018generalizing}, we report the average performance of all unseen domains. For {\it CIFAR-10-C}~\cite{hendrycks2019benchmarking}, we report the average performance of all types of corruptions at the highest level of severity.

{\bf Uncertainty assessment.}
We visualize feature perturbation $|\mathbf{e}|= |\mathbf{h}^+ -\mathbf{h}|$ and the embedding of domains at different training iterations $T$ on MNIST~\cite{lecun1998gradient}. 
We use t-SNE~\cite{maaten2008tsne} to visualize the source and augmented domains without and with uncertainty assessment in the embedding space.
Results are shown in Fig.~\ref{fig:feature}.
In the model  without  uncertainty (left), the feature perturbation $\mathbf{e}$ is sampled from $\mathcal{N}(\mathbf{0},\mathbf{I})$ without learnable parameters. In the model  with uncertainty (right),
we observe that most perturbations are located in the background area which increases the variation of $\mathcal{S}^+$ while keeping the category unchanged.
As a result, models with uncertainty can create large domain transportation in a curriculum learning scheme, yielding safe augmentation and improved accuracy on unseen domains.
We visualize the density of $\mathbf{y}^+$ in Fig.~\ref{fig:y_plus}. As seen, models with uncertainty can  significantly augment the label space.

{\bf Variational feature perturbation.} 
We investigate different designs of feature perturbation:  
{\it 1) Random Gaussian:} the feature perturbation $\mathbf{e}$ is sampled from $\mathcal{N}(\mathbf{0},\mathbf{I})$ without learnable parameters. 
{\it 2) Deterministic perturbation:} we directly add the learned  $\boldsymbol{\mu}$ to $\mathbf{h}$ without sampling, yielding $\mathbf{h}^+ \leftarrow \mathbf{h} + \text{Softplus}(\boldsymbol{\mu})$. 
{\it 3) Random $\boldsymbol{\mu}$:} the feature perturbation $\mathbf{e}$ is sampled from $\mathcal{N}(\mathbf{0},\boldsymbol{\sigma})$, where $\boldsymbol{\mu} = \mathbf{0}$.
{\it 4) Random $\boldsymbol{\sigma}$:} $\mathbf{e}$ is sampled from $\mathcal{N}(\boldsymbol{\mu}, \mathbf{I})$, where $\boldsymbol{\sigma} =  \mathbf{I}$. Results on these different choices are shown in Tab.~\ref{tab:ablation_x}. 
As seen, \textit{Random Gaussian} yields the lowest accuracy on both datasets, indicating the necessity of learnable perturbations. 
\textit{Deterministic perturbation} is inferior to \textit{Random $\boldsymbol{\mu}$} and  \textit{Random $\boldsymbol{\sigma}$}, suggesting that sampling-based perturbation can effectively increase the domain capacity. Finally, either \textit{Random $\boldsymbol{\mu}$} or \textit{Random $\boldsymbol{\sigma}$} is slightly worse than the full model. We conclude that both learnable $\boldsymbol{\mu}$ and learnable $\boldsymbol{\sigma}$ contribute to the final performance.

{\bf Learnable label mixup.} We implement two variants of label {\it mixup}: 
{\it 1) Without mixup:} the model is trained without label augmentation.
{\it 2) Random mixup:} the mixup coefficient $\lambda$ is sampled from a fixed distribution $\operatorname{Beta}(1,1)$. 
Results on the two variants are reported in Tab.~\ref{tab:ablation_y}. We notice that \textit{Random mixup} achieves better performance than \textit{without mixup}. The results support our claim that label augmentation can further improve the model performance. The learnable mixup (full model) achieves the best results, suggesting that the proposed learning label mixup can create informative domain interpolations for robust learning.

\begin{table}[]
    \centering
   	\resizebox{.9\linewidth}{!}{
		\begin{tabular}{@{}lcc@{}}
			\toprule
			   & Digits~\cite{volpi2018generalizing} & CIFAR-10-C~\cite{hendrycks2019benchmarking} \\
			\hline
			\textbf{Full Model} & \textbf{61.3}$\pm$0.73 &  \textbf{70.2}$\pm$0.62\\
			\hline
			w/o mixup & 60.6$\pm$0.76  & 67.4$\pm$0.64 \\
			Random mixup & 60.9$\pm$1.10 & 69.4$\pm$0.58 \\
			\bottomrule
	    \end{tabular}}
        \caption{Ablation study of label mixup.}\label{tab:ablation_y}
\end{table}

{\bf Training strategy.} At last, we compare different training strategies.
{\it 1) Without adversarial training:} models are learned without adversarial training (Eq.~\ref{eq:ada}).
{\it 2) Without meta-learning:}
the source $\mathcal{S}$ and augmentations $\mathcal{S}^+$ are trained together without the meta-learning scheme.
{\it 3) Without minimizing $\phi_p$:} $\phi_p$ is not optimized in Eq.~\ref{eq:mc}. 
Results are reported in Tab.~\ref{tab:ablation_l}. 
The adversarial training contributes most to the improvements:  9.5\%  on {\it Digits} and 10.2\% on {\it CIFAR-10-C}. 
Meta-learning consistently improve the accuracy and reduce the deviation on both datasets.
We notice that the accuracy is slightly dropped without minimization of $\phi_p$, possibly due to the excessive accumulation of perturbations.

\begin{table}[]
    \centering
    \resizebox{0.9\linewidth}{!}{
		\begin{tabular}{@{}lcc@{}}
			\toprule
			   & Digits~\cite{volpi2018generalizing} & CIFAR-10-C~\cite{hendrycks2019benchmarking} \\
			\hline
			\textbf{Full Model} & \textbf{61.3}$\pm$0.73 &  \textbf{70.2}$\pm$0.62\\
			\hline
    		w/o adv. training & 51.8$\pm$0.71 & 60.0$\pm$0.55 \\
    		w/o meta-learning & 60.9$\pm$1.24 & 68.7$\pm$0.81 \\
			w/o minimizing $\phi_p$ & 60.6$\pm$0.91  & 69.6$\pm$0.75  \\
			\bottomrule
	    \end{tabular}}
    \caption{Ablation study of training strategy.}\label{tab:ablation_l}
\end{table}

\section{Conclusion}
In this work, we introduced uncertain out-of-domain generalization to tackle the problem of single source generalization. Our method explicitly model the uncertainty of domain augmentations in both input and label spaces. In input space, the proposed uncertainty-guided feature perturbation resolves the limitation of raw data augmentation, yielding a domain-knowledge-free solution for various modalities. In label space, the proposed uncertainty-guided label mixup further increases the domain capacity.
The proposed \textit{domain uncertainty score} is capable of estimating uncertainty with high efficiency, which plays a crucial role for safe deployment.
Finally, the proposed Bayesian meta-learning framework can maximize the posterior distribution of domain augmentations, such that the learned model can generalize well on unseen domains. The experimental results prove that our method achieves superior performance on a wide scope of tasks, including \textit{image classification}, \textit{semantic segmentation}, \textit{text classification}, and \textit{speech recognition}. 
In addition to the superior performances we achieved through these experiments, a series of ablation studies further validate the effectiveness of key components in our method. In the future, we expect to extend our work to semi-supervised learning or knowledge transferring in multimodal learning.


{\small
\bibliographystyle{ieee_fullname}
\bibliography{egbib, egbib_peng}
}

\clearpage
\begin{appendices}

\begin{figure*}[t]
\begin{center}
\subfigure{
	\includegraphics[width=0.18\linewidth]{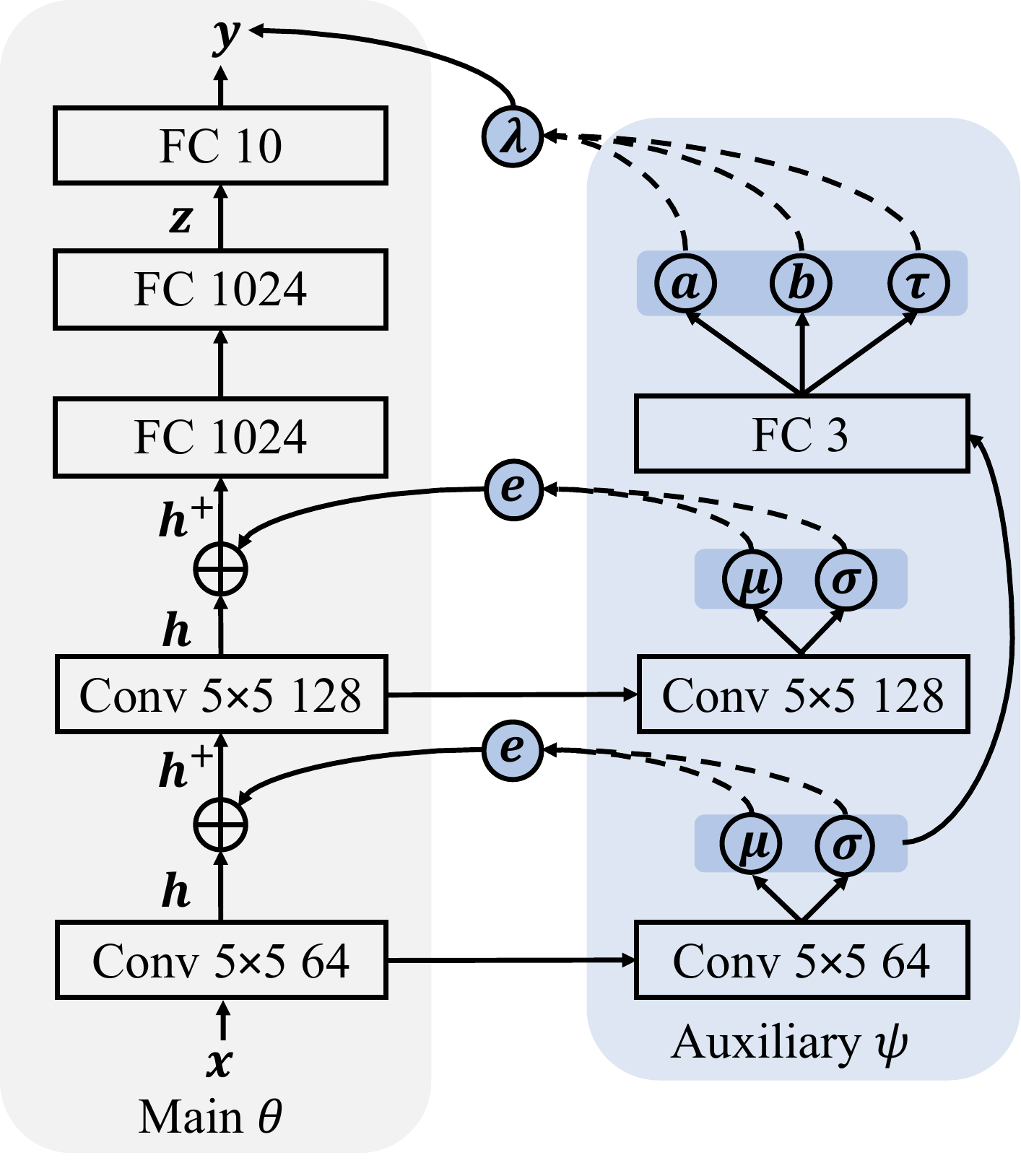}
}
\subfigure{
	\includegraphics[width=0.18\linewidth]{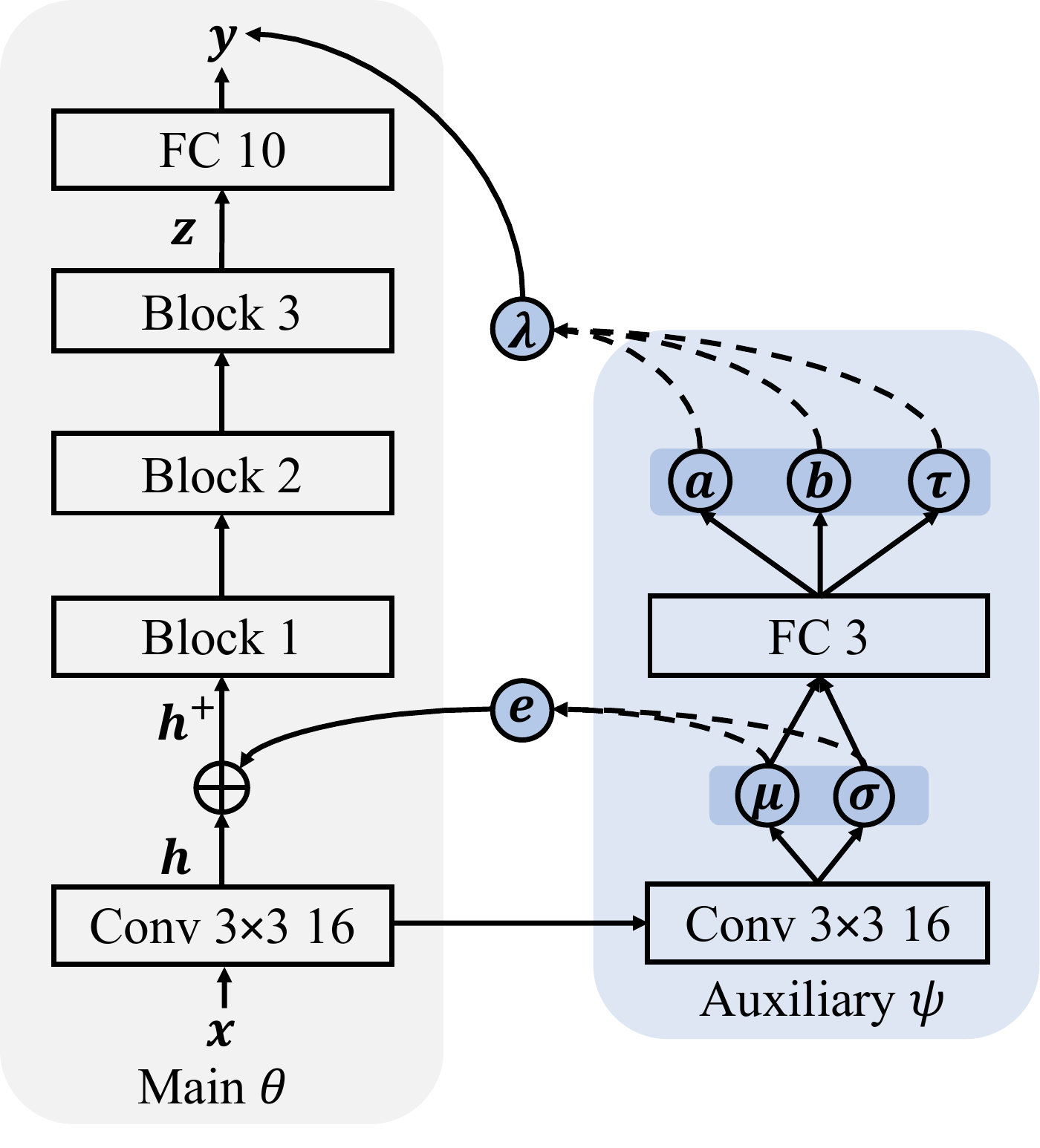}
}
\subfigure{
	\includegraphics[width=0.18\linewidth]{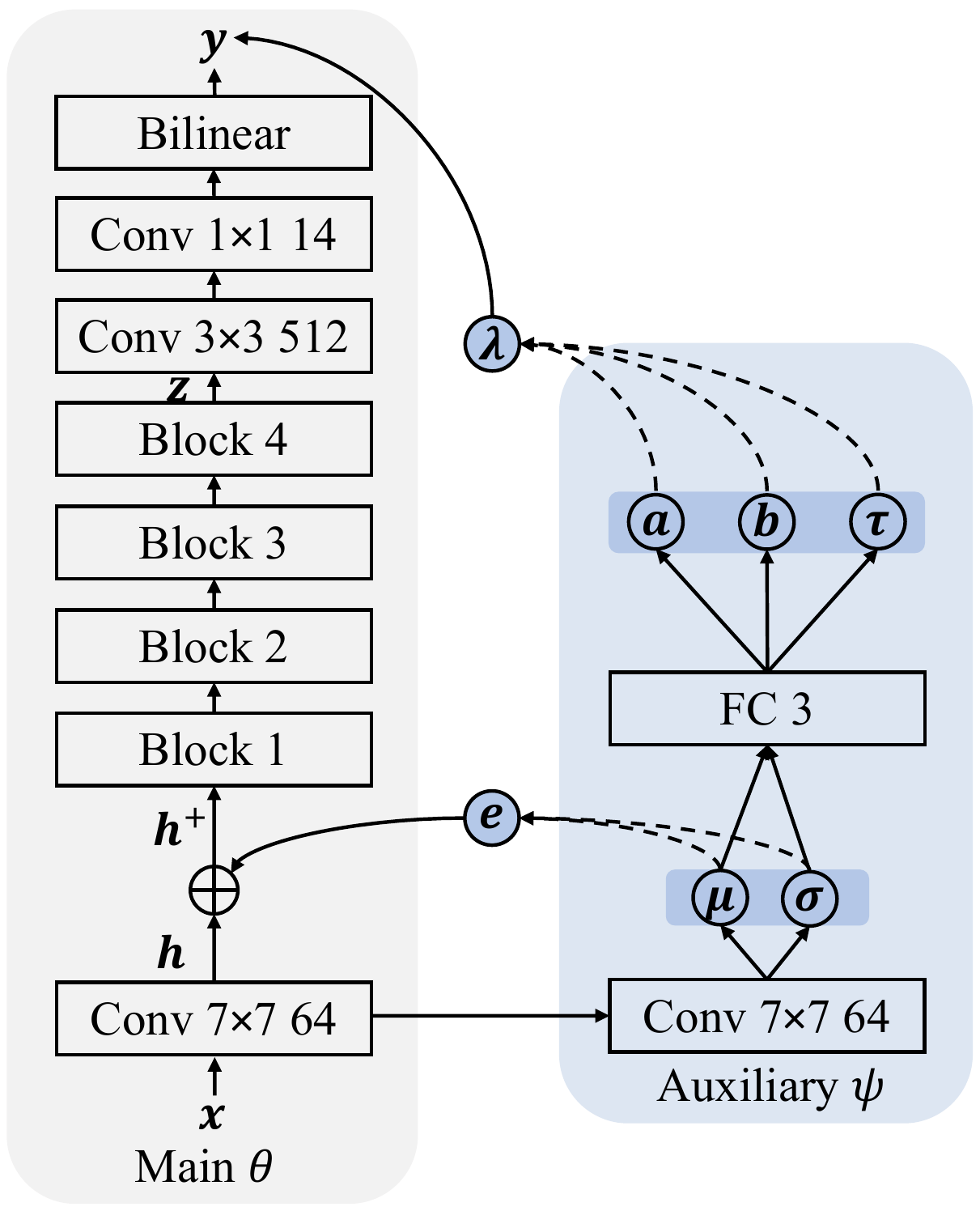}
}
\subfigure{
	\includegraphics[width=0.18\linewidth]{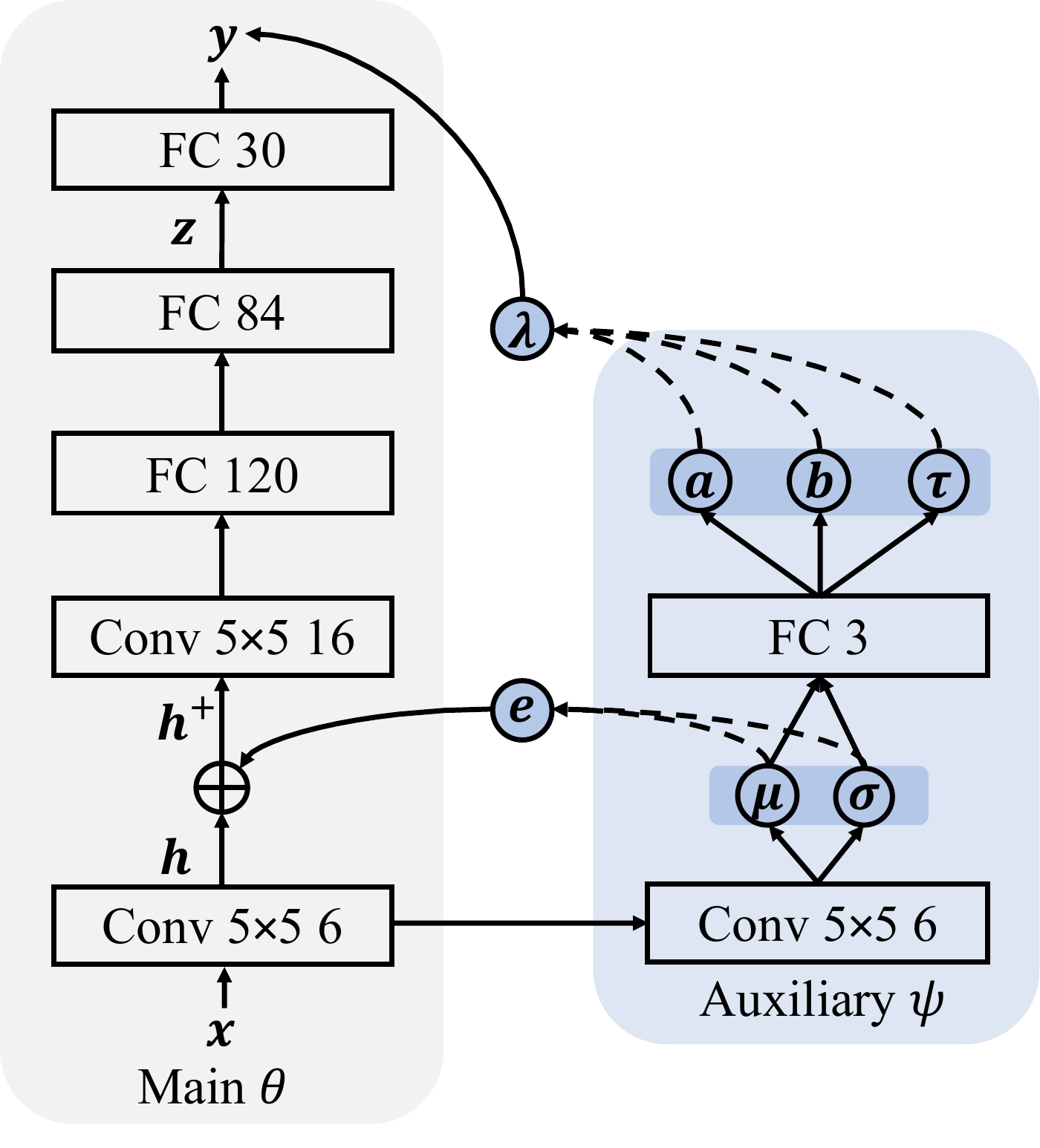}
}
\subfigure{
	\includegraphics[width=0.18\linewidth]{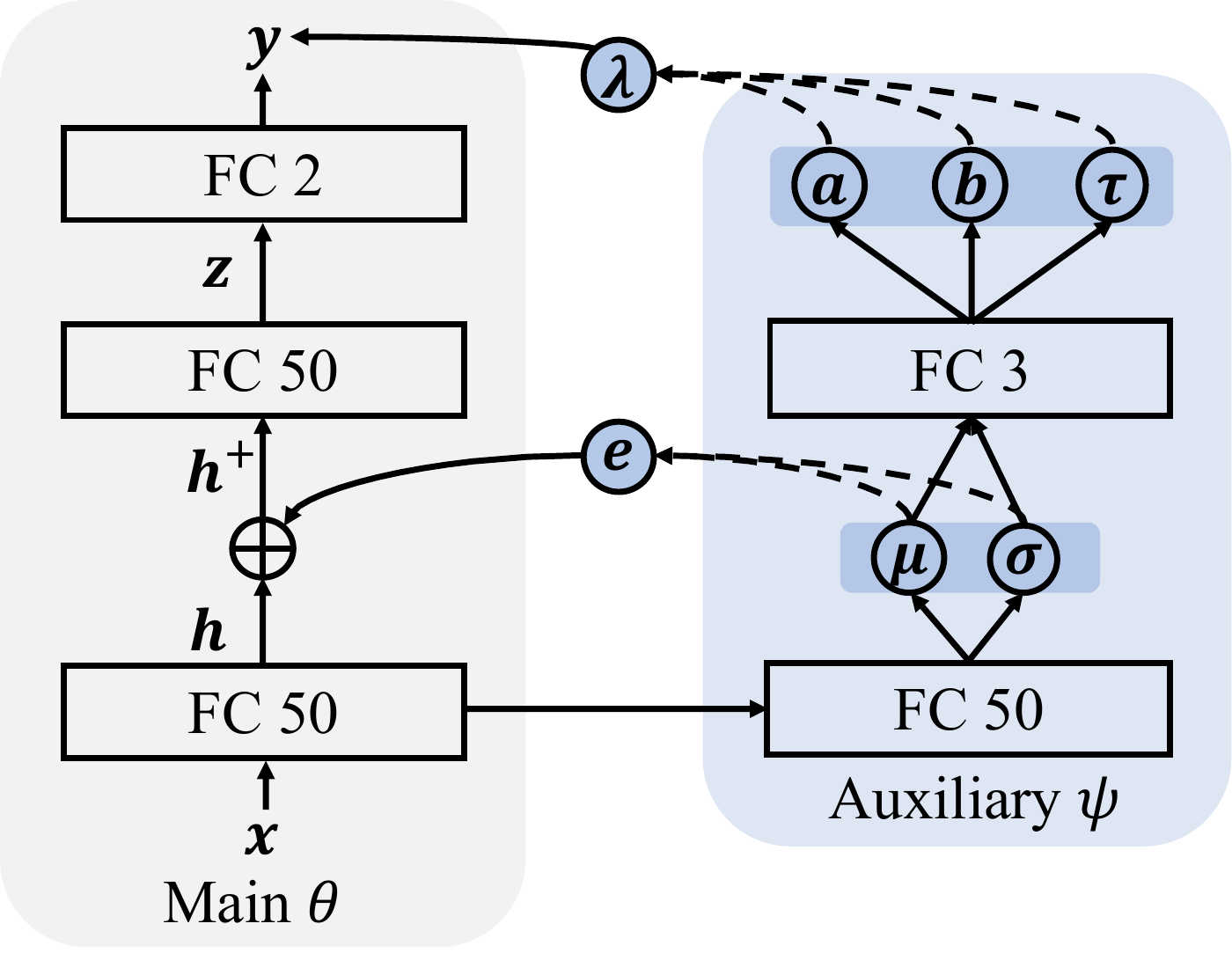}
}
\end{center}
\vspace{-1em}
\caption{Architectures of main and auxiliary models. \textbf{From left to right:} (a) {\it Digits}~\cite{volpi2018generalizing}; (b) \textit{CIFAR-10-C}~\cite{hendrycks2019benchmarking}; (c) {\it SYTHIA}~\cite{ros2016synthia}; (d) {\it Google Commands}~\cite{warden2018speech}, and (e) {\it Amazon Reviews}~\cite{chen2012marginalized}.} 
\label{models}
\end{figure*}

\begin{figure*}
\centering
\begin{minipage}[h]{0.62\textwidth}
\centering
\includegraphics[width=1.0\linewidth]{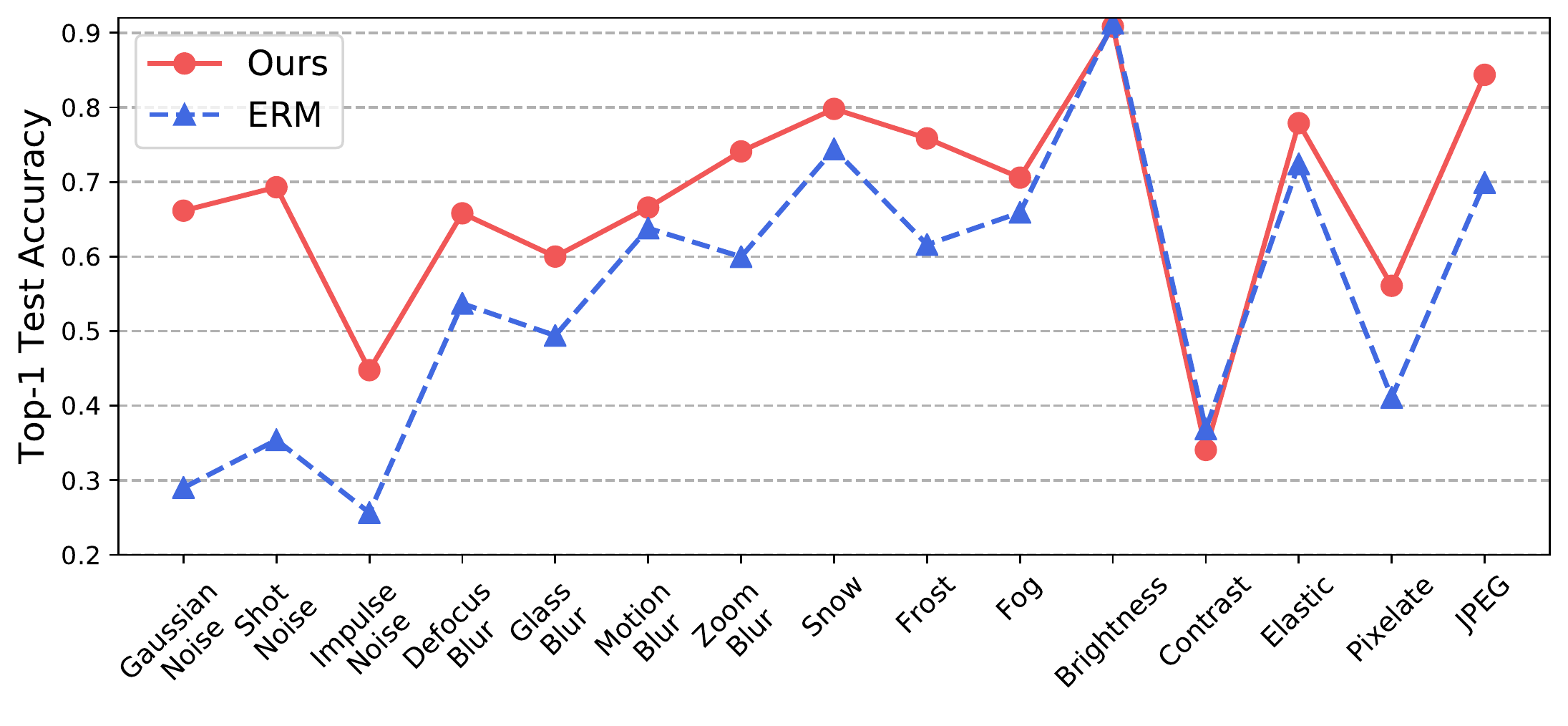}
\vspace{-.7em}
\caption{Classification accuracy on fifteen corruptions of \textit{CIFAR-10-C} using the backbone of WRN (40-2). Following Fig.~\ref{fig:cifar}, the accuracy of each corruption with the highest level of severity is presented. Our method achieves 20\% improvements on corruptions of noise.}
\label{fig:cifar5}
\end{minipage}
\hspace{0.1in}
\begin{minipage}{0.34\textwidth}
\vspace{-0.3in}
\centering
\includegraphics[width=1.0\linewidth]{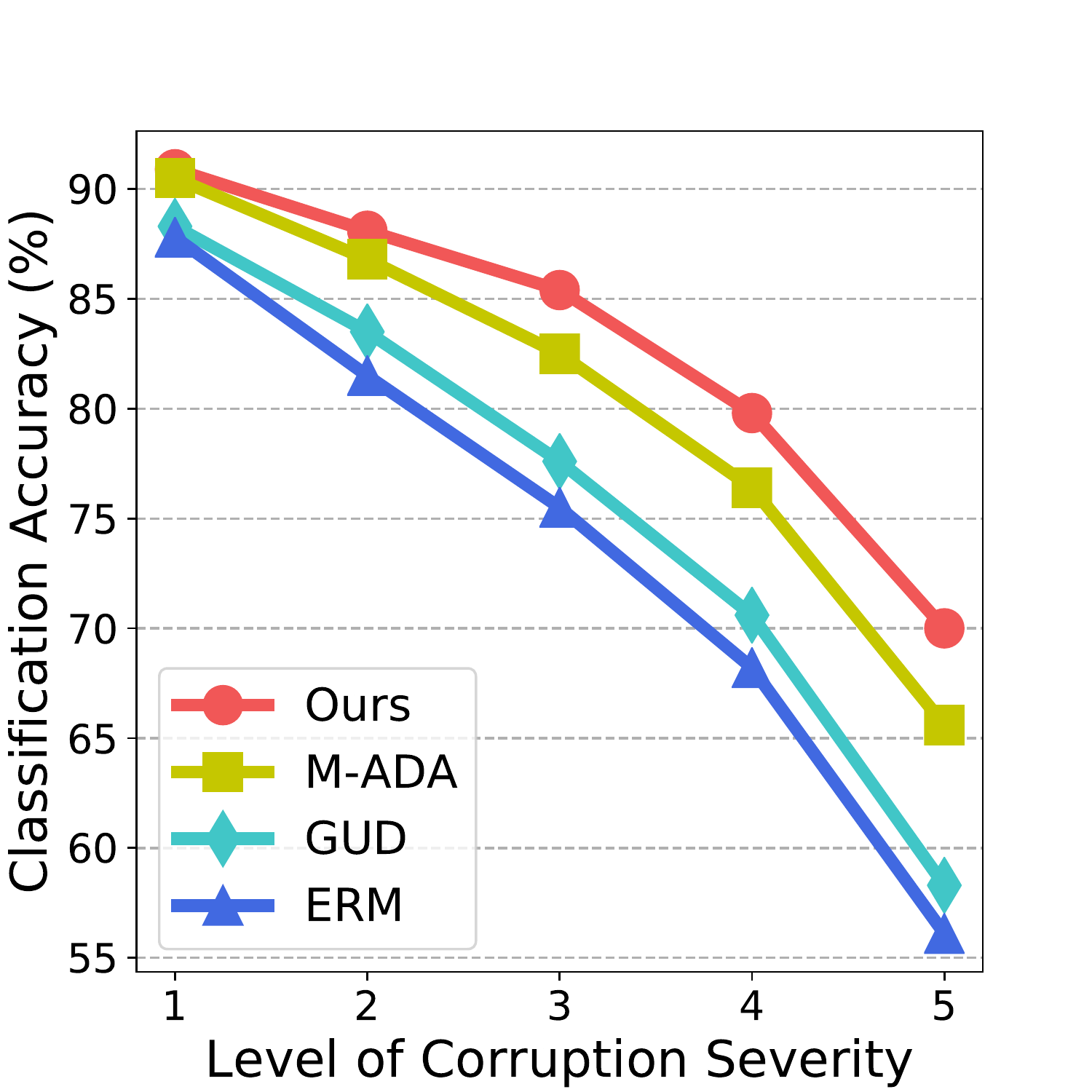}
\vspace{-0.23in}
\caption{Classification accuracy (\%) on five levels of corruption severity. Our method has the smallest degradation under the highest level of corruption severity.} \label{fig:cifar}
\end{minipage}
\vspace{-1.5em}
\end{figure*}

\section{Architecture Design and Setup}\label{sec:setup}

We provide more experimental details on the five datasets: {\it Digits}~\cite{volpi2018generalizing}, 
{\it CIFAR-10-C}~\cite{hendrycks2019benchmarking}, {\it SYTHIA}~\cite{ros2016synthia}, {\it Amazon Reviews}~\cite{chen2012marginalized}, and {\it Google Commands}~\cite{warden2018speech}. 
In learnable label mixup, we use Gaussian parameters of feature perturbations from the first layer.
We choose specific backbone models, and design different auxiliary models as well as training strategies according to characteristics of each dataset.

In {\it Digits}~\cite{volpi2018generalizing}, the backbone model is \textit{conv-pool-conv-pool-fc-fc-softmax}.
There are two 5 $\times$ 5 convolutional layers with 64 and 128 channels respectively. Each convolutional layer is followed by a max pooling layer with the size of 2 $\times$ 2. The size of the two Fully-Connected (FC) layers is 1024 and the size of the softmax layer is 10. We inject perturbations to latent features of the two convolutional layers. The detailed architecture is presented in Fig.~\ref{models} (a). We employ Adam~\cite{kingma2014adam} for optimization with batch size of 32. We train for total 10K iterations with learning rate of $10^{-4}$.

In \textit{CIFAR-10-C}~\cite{hendrycks2019benchmarking}, we evaluate our method on two backbones: AllConvNet (AllConv)~\cite{salimans2016weight} and Wide Residual Network (WRN)~\cite{zagoruyko2016wide} with 40 layers and the width of 2. In AllConv~\cite{salimans2016weight}, the model starts with three 3 $\times$ 3 convolutional layers with 96 channels. Each layer is followed by batch normalization (BN)~\cite{ioffe2015batch} and GELU. They convert the original image with three channels to feature maps of 96 channels. Then, the features go though three 3 $\times$ 3 convolutional layers with 192 channels. After that, the features are fed into two 1 $\times$ 1 convolutional layers with 192 channels and an average pooling layer with the size of 8 $\times$ 8. Finally, a softmax layer with the size of 10 is used for classification. 
In WRN~\cite{zagoruyko2016wide}. The first layer is a 3$\times$3 convolutional layer. It converts the original image with three channels to feature maps of 16 channels. Then the features go through three blocks of 3$\times$3 convolutional layers. Each block consists of six basic blocks and each basic block is composed of two convolutional layers with the same number of channels. And their channels are \{32, 64, 128\} respectively. Each layer is followed by batch normalization (BN)~\cite{ioffe2015batch}. An average pooling layer with the size of 8 $\times$ 8 is appended to the output of the third block. Finally, a softmax layer with the size of 10 is used for prediction. In both AllConv~\cite{salimans2016weight} and WRN~\cite{zagoruyko2016wide}, we only inject perturbations to the latent features of the first convolutional layer. We also tried to inject perturbations in the next few layers or blocks, however, we found the performance degraded severely mainly due to its large effect on the semantic feature, {\it i.e.}, outputs before the activation layer. The detailed architecture with backbone of WRN is shown in Fig.~\ref{models} (b). Following the training procedure in \cite{zagoruyko2016wide}, we use SGD with Nesterov momentum and set the batch size to 128. The initial learning rate is 0.1 with a linear decay and the number of epochs is 200.

In {\it SYTHIA}~\cite{ros2016synthia}, we use FCN-32s~\cite{long2015fully} with the backbone of ResNet-50~\cite{he2016deep}. The model consists of a feature extractor and a classifier. We use ResNet-50~\cite{he2016deep} as the feature extractor, which is composed of a 7$\times$7 convolutional layer with 64 channels and four convolutional blocks. The classifier consists of a 
3$\times$3 convolutional layer with 512 channels, a 
1$\times$1 convolutional layer with 14 channels, and a bilinear layer used to up-sample the coarse outputs to the original size. We use Adam with the learning rate $\alpha=0.0001$. We set the batch size to 8 and the number of epochs to 50.

In {\it Amazon Reviews}~\cite{chen2012marginalized},
reviews are assigned binary labels - 0 if the rating of the product is up to 3 stars, and 1 if the rating is 4 or 5 stars. The extracted features are fed into two FC layers with the size of 50. A softmax layer with the size of two is used to classify the sentiment of reviews into ``positive'' or   ``negative''. 
All models are trained using  Adam~\cite{kingma2014adam} optimizer with learning rate of $10^{-4}$ and batch size of 32 for 1000 iterations.
In {\it Google Commands}~\cite{warden2018speech}, 
the mel-spectrogram features are fed into LeNet~\cite{Lecun98gradient} as one-channel input. 
The original image is fed into two 5 $\times$ 5 convolutional layers with the channels of 6 and 16, respectively. Next, the features go through two FC layers with the size of 120 and 84, respectively. Finally, a softmax layer with the size of 30 is leveraged to predict the spoken word. 
Models are trained using Adam~\cite{kingma2014adam} with learning rate $10^{-4}$ and batch size of 128 for 30 epoches.
For the corrupted test sets, the range of ``amplitude change'' is (0.7, 1.1). The maximum scales of ``pitch change'', ``background noise'', and ``stretch'' are 0.2, 0.45, and 0.2, respectively. The maximum shift of ``time shift'' is 8.
In the experiments of {\it Amazon Reviews}~\cite{chen2012marginalized} and {\it Google Commands}~\cite{warden2018speech}, feature perturbations are appended to the first layer. The detailed architectures used for {\it Google Commands}~\cite{warden2018speech} and  {\it Amazon Reviews}~\cite{chen2012marginalized} are presented in Fig.~\ref{models} (c) and (d), respectively.

\section{Additional Results}\label{sec:ad_results}

\subsection{Image Classification}

{\bf 1) Classification accuracy on \textit{CIFAR-10-C}}~\cite{hendrycks2019benchmarking}. We train all models on clean data, {\it i.e.}, CIFAR-10, and test them on corruption data, {\it i.e., CIFAR-10-C}. In this case, there are totally $15$ unseen testing domains. We compare our method with the other three methods for domain generalization: ERM~\cite{koltchinskii2011oracle}, GUD~\cite{volpi2018generalizing}, and M-ADA~\cite{qiao2020learning}. The classification results on corruptions across five levels of severity are shown in Fig.~\ref{fig:cifar}. As seen, our method outperforms other methods across all levels of corruption severity. Specifically, the  gap  between  M-ADA~\cite{qiao2020learning} (previous SOTA) and our method gets larger with the level of severity increasing.
Fig.~\ref{fig:cifar5} shows more detailed comparison of all corruptions at the highest level of severity. As seen, our method achieves substantial gains across a wide variety of corruptions, with a small drop of performance in only two corruption types: brightness and contrast. Especially, accuracy is significantly improved by 20\% on corruptions of noise. Results demonstrate its strong generalization capability on severe corruptions.

\begin{table}[t]
\centering
		\resizebox{1.0\linewidth}{!}{
		\begin{tabular}[t]{@{}lccccc@{}}
			\toprule
			Method & $\lvert  \mathcal{T} \rvert$  & U \ $\rightarrow$ M & M \ $\rightarrow$ S &  S \ $\rightarrow$ M & Avg.\\
			\hline
			DIRT-T~\cite{shu2018dirt} & \multirow{3}{*}{All}&   - &54.50 & \textbf{99.40}&-\\
			SE~\cite{french2017self} & &    \textbf{98.07} & 13.96 &99.18& 70.40\\
			SBADA~\cite{russo2018source} & &   97.60 & \textbf{61.08} &76.14& 78.27\\
			\hline
			FADA~\cite{motiian2017few}  & 7  & 91.50& 47.00 & 87.20&  75.23\\
			CCSA~\cite{motiian2017unified}  & 10  & 95.71 &37.63 & 94.57 & 75.97\\
			\hline
			\hline
			\multirow{2}{*}{\bf Ours}
			&7 & 92.97  & 58.12  & 89.30 & 80.13\\
			&10 & 93.16 & 59.77  & 91.67 & \textbf{81.53} \\
			\bottomrule
		\end{tabular}}
\captionof{table}{Few-shot domain adaptation accuracy (\%) on {\it MNIST(M), USPS(U), and SVHN(S)}. $\lvert \mathcal{T} \rvert$ denotes the number of target samples (per class) used during model training.}\label{tab:few}
\end{table}

{\bf 2) Few-shot domain adaptation.}
We conduct three few-shot domain adaption tasks: \textit{USPS(U)}$\rightarrow$\textit{MNIST(M)}, \textit{MNIST(M)}$\rightarrow$\textit{SVHN(S)}, and \textit{SVHN(S)}$\rightarrow$\textit{MNIST(M)}.
Results of the three tasks are shown in Tab.~\ref{tab:few}. As seen, the result on the hardest
task (\textit{M}$\rightarrow$\textit{S}) is even competitive to that of SBADA~\cite{russo2018source} which requires all images of the target domain during training.
Specifically, our method achieves the best performance on the average of the three tasks. Note that, when the target domain changes, our method only needs to fine-tune the pre-trained model with a few samples within a small number of iterations, while other methods have to train entirely new models.

\subsection{Semantic Segmentation}

\begin{figure}[t]
\begin{center}
\includegraphics[width=1\linewidth]{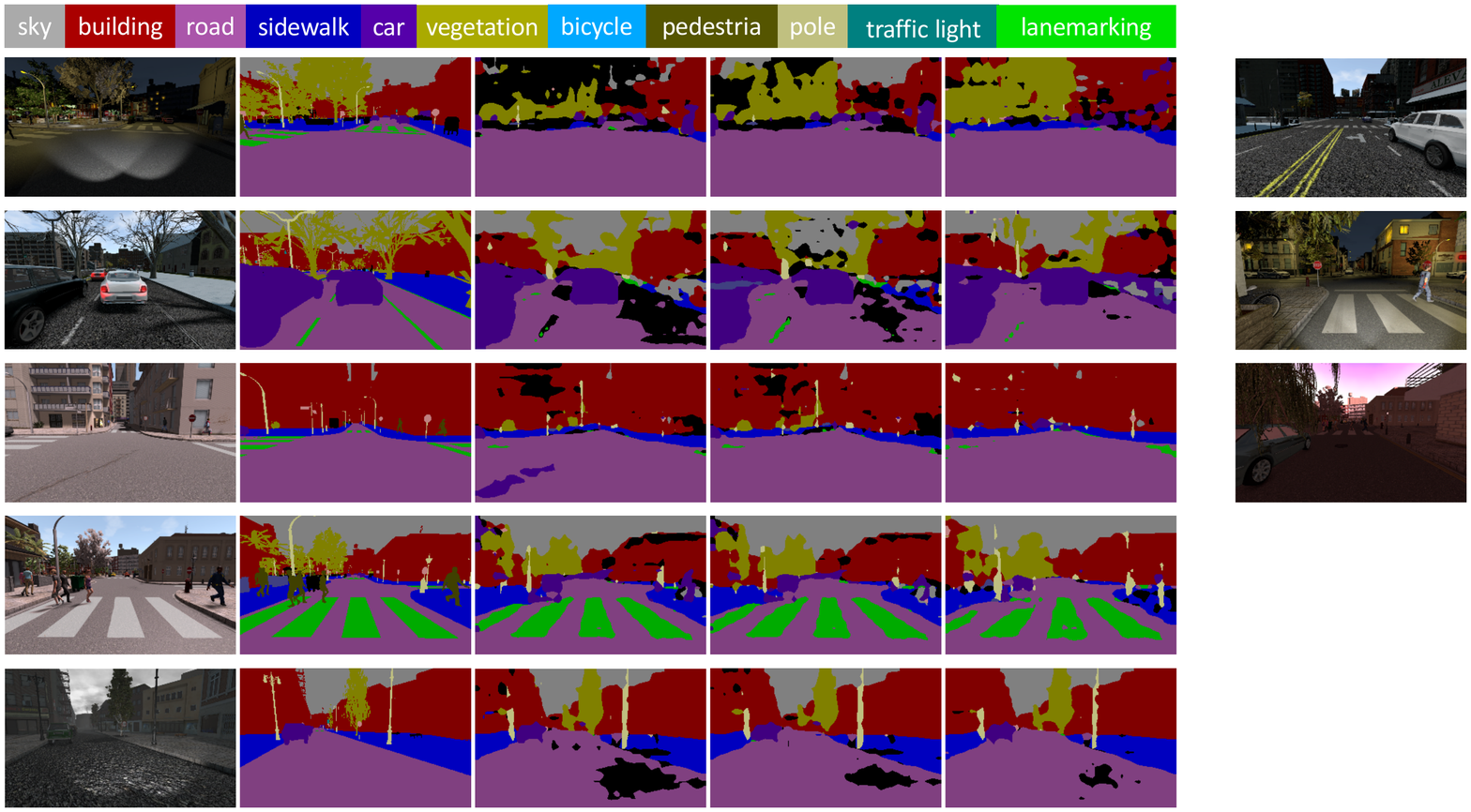}
\end{center}
\caption{Examples of semantic segmentation on {\it SYTHIA}~\cite{ros2016synthia}. {\bf From left to right:} (a) images from unseen domains; (b) ground truth; (c) results of ERM~\cite{koltchinskii2011oracle}; (d) results of M-ADA~\cite{qiao2020learning}; and (e) results of our method. Best viewed in color and zoom in for details.} 
\label{fig:seg_vis}
\end{figure}

In the experiment on {\it SYTHIA}~\cite{ros2016synthia}, \textit{Highway} is the source domain, and \textit{New York-like City} together with \textit{Old European Town} are unseen domains. 
Visual comparison on {\it SYTHIA}~\cite{ros2016synthia} is shown in Fig.~\ref{fig:seg_vis}.
Results demonstrate that our model can better generalize to the changes of locations, weather, and time.

\begin{table*}[h]
\begin{center}
\resizebox{.9\linewidth}{!}{
\begin{tabular}{@{}lcccccccccccc@{}}
	\toprule
	       & \multicolumn{3}{c}{books} & \multicolumn{3}{c}{dvd} & \multicolumn{3}{c}{kitchen} & \multicolumn{3}{c}{electronics}\\
	\cmidrule(lr){2-4} \cmidrule(lr){5-7} \cmidrule(lr){8-10} \cmidrule(lr){11-13}
	Method    & d & k & e & b & k & e & b & d & e & b & d & k  \\
	\hline
	ERM~\cite{koltchinskii2011oracle}    & 78.7  & 74.6 & 63.6 & 78.5 & 82.1 & \textbf{75.2} & \underline{75.4 } & 76.0 & 81.2 & \underline{69.4}  & \textbf{74.8} & 83.9 \\
	GUD~\cite{volpi2018generalizing}   & 79.1 & 75.6 &  64.7 & 78.1 & 82.0 & 74.6 & 74.9  & 76.7  & 81.6 & 68.9 & 74.2 & 84.4 \\
	M-ADA~\cite{qiao2020learning}   & \underline{79.4} & \underline{76.1} & \underline{65.3}  & \underline{78.8} & \underline{82.6}  & 74.3 &  75.2 & \underline{77.3} & \underline{82.3} & 69.0 & 73.7 & \underline{84.8}\\
	\hline
    \textbf{Ours}     & \textbf{80.2}  & \textbf{76.8} &\textbf{67.1} & \textbf{80.1} & \textbf{83.5} & \underline{75.0} & \textbf{76.1}   &  \textbf{78.2}&  \textbf{83.5} & \textbf{70.2} & \underline{74.5}  & \textbf{85.7} \\
	\bottomrule
\end{tabular}}
\captionof{table}{Text classification accuracy (\%) on {\it Amazon Reviews}~\cite{chen2012marginalized}. The models are trained on only one text domain and evaluated on other unseen text domains. Our method outperforms others in all settings except ``$dvd \rightarrow electronics$'' and ``$electronics \rightarrow dvd$''. The possible reason is that ``$dvd$'' and ``$electronics$'' may share a similar distribution while our method creates large distribution shift.}\label{tab:text}
\end{center}
\end{table*}

\subsection{Text Classification}
We train the models on one source domain, and evaluate them on the other three domains. 
Tab.~\ref{tab:text} shows the results of text classification on {\it Amazon Reviews}~\cite{chen2012marginalized} . We found that our method outperform previous ones on all the three unseen domains when the source domain is ``books'' or ``kitchen''. Specially, our method outperforms ERM~\cite{koltchinskii2011oracle} by 3.5\% on ``books $\rightarrow$ electronics''.
We observe that there is a little drop in accuracy on ``dvd $\rightarrow$ electronics'' and ``electronics $\rightarrow$ dvd''.
One possible reason is that ``electronics'' and ``dvd'' may share a similar distribution. And our method creates large distribution shift, degrading the performance on them.

\subsection{Ablation Study}
\begin{figure}[h]
\vspace{-0.2in}
\begin{center}
\subfigure{
	\includegraphics[width=0.47\linewidth]{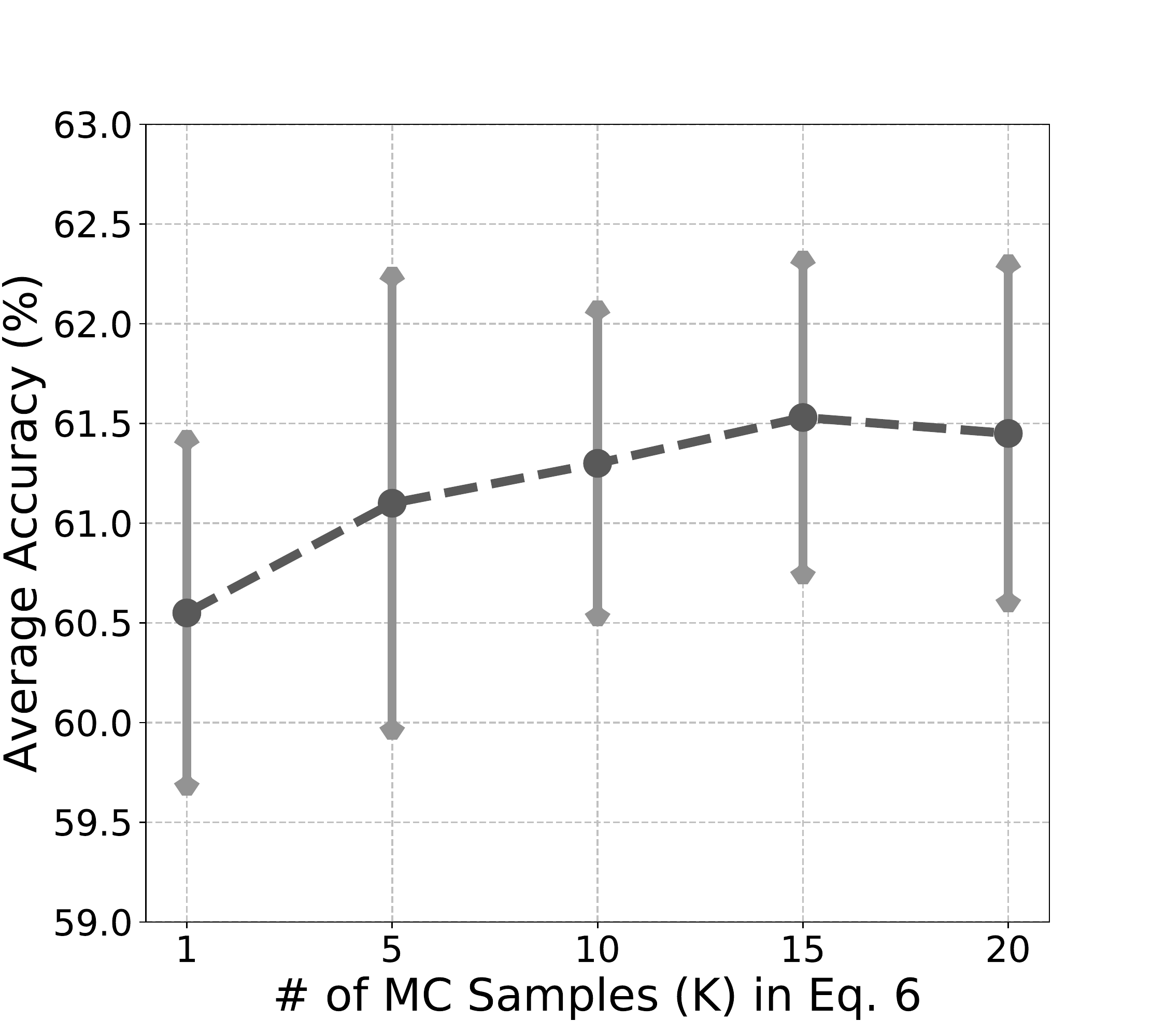}
}
\subfigure{
	\includegraphics[width=0.47\linewidth]{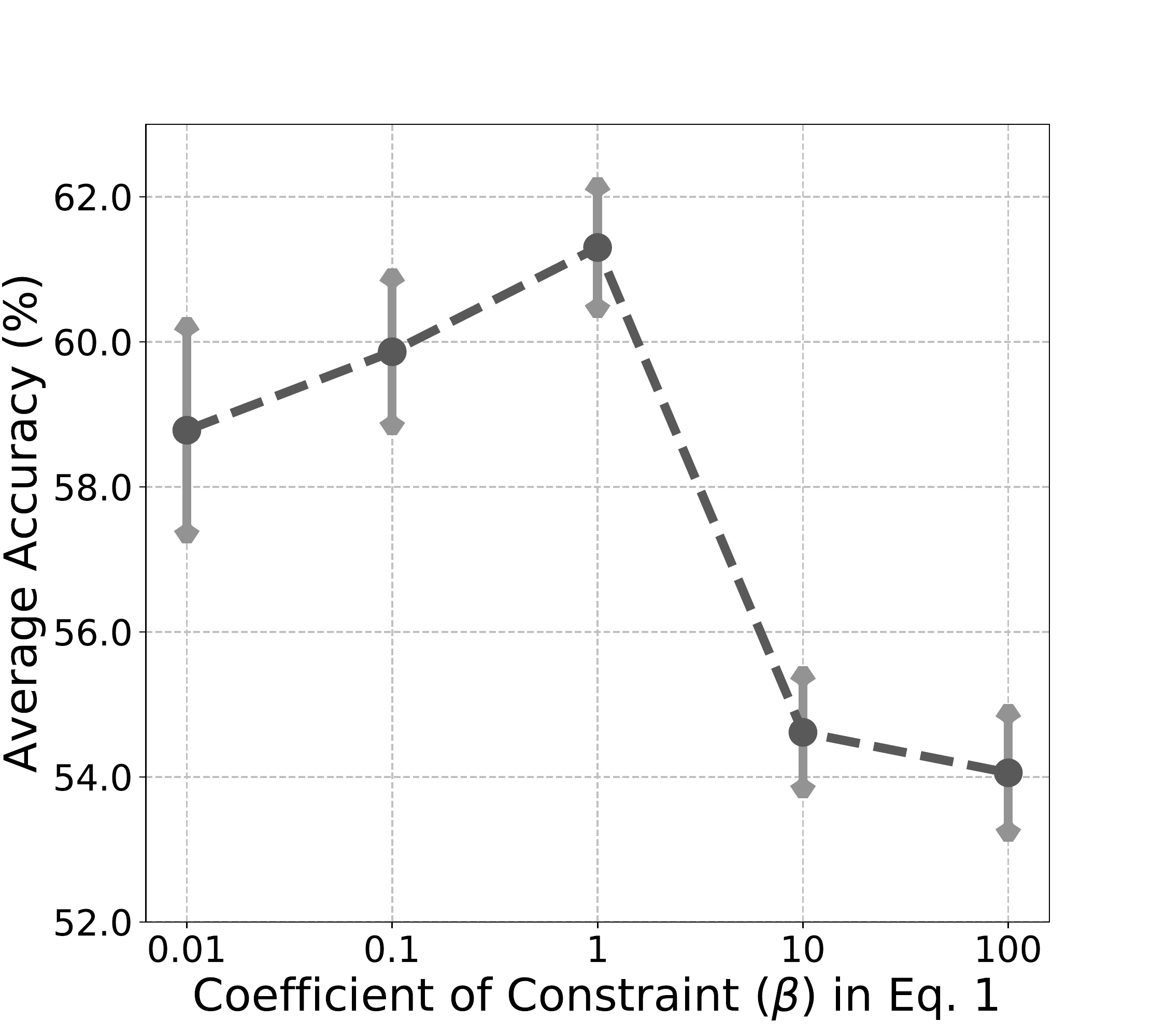}
}
\end{center}
\vspace{-1em}
\caption{Ablation study on hyper-parameters $K$ and $\beta$.
The average accuracy on the four unseen domains (\textit{MNIST-M}~\cite{ganin2015unsupervised}, \textit{SVHN}~\cite{netzer2011reading}, \textit{SYN}~\cite{ganin2015unsupervised}, and \textit{USPS}~\cite{denker1989advances}) is presented. We set $K=15$ and  $\beta=1$ according to the best classification accuracy.} 
\label{fig:params}
\end{figure}

We study the effect of two important hyper-parameters of our model: the number of Monte-Carlo (MC) samples ($K$) and the coefficient of constraint ($\beta$).
We report the average accuracy on the four unseen domains (\textit{MNIST-M}~\cite{ganin2015unsupervised}, \textit{SVHN}~\cite{netzer2011reading}, \textit{SYN}~\cite{ganin2015unsupervised}, and \textit{USPS}~\cite{denker1989advances}).
We present the classification results under different hyper-parameters in Fig.~\ref{fig:params}.

{\bf 1) Number of MC samples ($K$).} The classification accuracy on {\it Digits}~\cite{volpi2018generalizing} with different $K$ is shown in Fig.~\ref{fig:params} (a).
We notice that the average accuracy gradually increases from $K = 1$ to $K = 15$ and remains stable when $K = 20$.

{\bf 2) Coefficient of constraint ($\beta$).} The constraint is used to make adversarial domain augmentation satisfy the worst-case constraint.
Results on {\it Digits}~\cite{volpi2018generalizing} with different $\beta$ is presented in Fig.~\ref{fig:params} (b). As seen, the accuracy falls dramatically when $\beta = 10$, because large $\beta$ may severely limit the domain transportation and create domain augmentations similar to the source.

\end{appendices}

\end{document}